\documentclass{article} 
\usepackage[final]{colm2026_conference}

\usepackage{microtype}
\usepackage{hyperref}
\usepackage{url}
\usepackage{booktabs}

\usepackage{lineno}
\usepackage{graphicx}    
\graphicspath{{media/}} 
\usepackage{enumitem}
\usepackage{booktabs}
\usepackage{makecell}
\usepackage{multirow}
\usepackage{tabularx}
\usepackage{xcolor}
\newcommand{\fan}[1]{\textcolor{red}{#1}}
\newcommand{\gee}[1]{\textcolor{blue}{#1}}
\newcommand{\orange}[1]{\textcolor{orange}{#1}}
\usepackage{float}
\usepackage{amsmath}
\usepackage{wrapfig}
\usepackage{tcolorbox}
\definecolor{darkblue}{rgb}{0, 0, 0.5}
\hypersetup{colorlinks=true, citecolor=darkblue, linkcolor=darkblue, urlcolor=darkblue}

\title{Train Yourself as an LLM: Exploring Effects of AI Literacy on Persuasion via Role-playing LLM Training}

\author{
Qihui Fan$^{1}$, Min Ge$^{2}$, Chenyan Jia$^{1,2}$, Weiyan Shi$^{1}$ \\
$^{1}$Khoury College of Computer Sciences, Northeastern University \\
$^{2}$College of Arts, Media and Design, Northeastern University \\
\texttt{\{fan.qih, ge.min, c.jia, we.shi\}@northeastern.edu}
}

\begin{document}

\ifcolmsubmission
\linenumbers
\fi

\maketitle

\begin{abstract}
As large language models (LLMs) become increasingly persuasive, there is concern that people’s opinions and decisions may be influenced across various contexts at scale. 
Prior mitigation (e.g., AI detectors and disclaimers) largely treats people as passive recipients of AI-generated information. To provide a more proactive intervention against persuasive AI, we introduce \textbf{LLMimic}\footnote{
Project: \url{https://llmimic.lifilmstudio.com}; Code \& Data: \url{https://github.com/CHATS-lab/LLMimic}
}, a role-play-based, interactive, gamified AI literacy tutorial, where participants assume the role of an LLM and progress through three key stages of the training pipeline (pretraining, SFT, and RLHF). We conducted a $2 \times 3$ between-subjects study ($N = 274$) where participants either (1) watched an AI history video (control) or (2) interacted with LLMimic (treatment), and then engaged in one of three realistic AI persuasion scenarios: (a) charity donation persuasion, (b) malicious money solicitation, or (c) hotel recommendation. Our results show that LLMimic significantly improved participants’ AI literacy ($p < .001$), reduced persuasion success across scenarios ($p < .05$), and enhanced truthfulness and social responsibility levels ($p<0.01$) in the hotel scenario. These findings suggest that LLMimic offers a scalable, human-centered approach to improving AI literacy and supporting more informed interactions with persuasive AI.
\end{abstract}
\vspace{-0.8em}
\begin{figure}[H]
  \centering
  \includegraphics[width=0.85\linewidth]{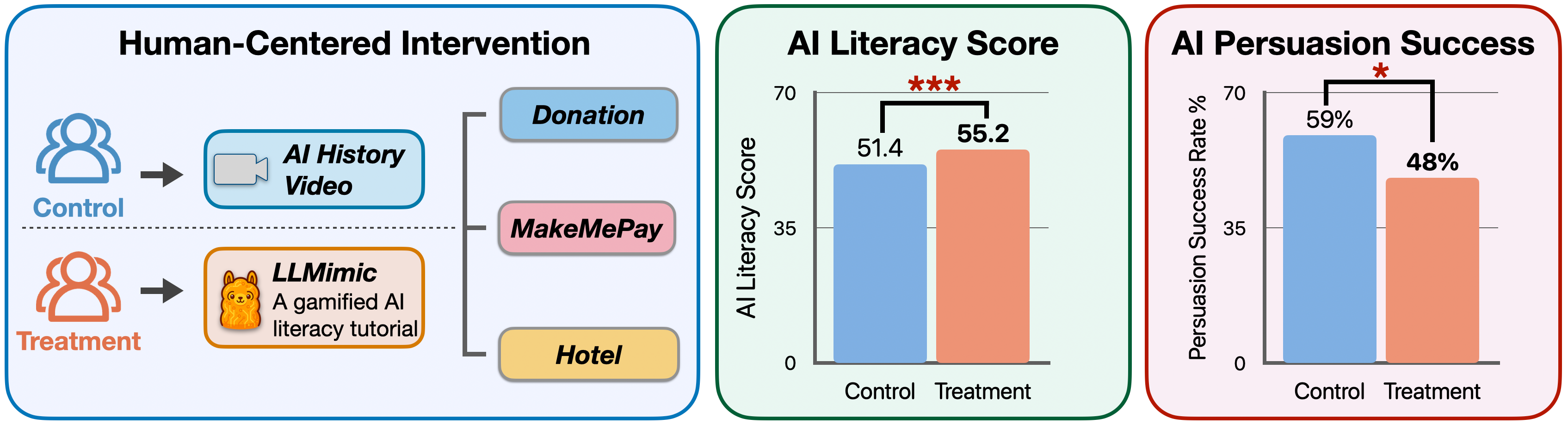}
    \caption{We developed LLMimic, a role-play-based, interactive, gamified AI literacy tutorial, and conducted a 2 (Intervention: AI history video vs.\ LLMimic) $\times$ 3 (Persuasion Scenarios: Donation, MakeMePay, Hotel recommendation) between-subjects human study. The results show that LLMimic significantly improves people’s AI literacy and reduces the AI persuasion success rate across scenarios, serving as  an effective mitigation.}
\label{fig:teaser}
\end{figure}

\section{Introduction}
Large language models (LLMs) have greatly enhanced the persuasive capabilities of AI systems. Besides early empirical evidence \citep{wang2020persuasiongoodpersonalizedpersuasive}, recent work shows that LLMs can produce credible, scalable, and personalized persuasion across domains for prosocial purposes such as pro-vaccination \citep{10.1145/3579592} and longitudinal conspiracy belief reduction \citep{doi:10.1126/science.adq1814}. However, LLMs may also produce biased content or misinformation \citep{biased_AI_write_assi,doi:10.1073/pnas.2322823121} and sway opinions in different directions \citep{costello2026large}, which can mislead users’ judgments and cause harm. 

Prior work has explored various methods to mitigate AI persuasion, such as detecting persuasive content \citep{wang2024mentalmanipdatasetfinegrainedanalysis,modzelewski2026aigeneratedpersuasiondetectedpersuaficial}. However, these approaches suffer from inconsistent detection accuracy: while overt AI-generated persuasion is relatively easy to identify \citep{modzelewski2026aigeneratedpersuasiondetectedpersuaficial}, recognizing subtle persuasion cues remains challenging \citep{modzelewski2026aigeneratedpersuasiondetectedpersuaficial,srivastav2026unknownunknownshiddenintentions}. Moreover, persuasion effects can persist even when messages are labeled as AI-generated or potentially biased \citep{gallegos2025labelingmessagesaigenerateddoes,biased_AI_write_assi}. Critically, these detection-based methods largely treat users as passive recipients rather than empowering them to discern persuasive intent and critically evaluate persuasive content. To address this gap, we explore more proactive, human-centered approaches focused on improving people's AI literacy.

Studies have shown that AI literacy enables people to more critically evaluate AI-generated content \citep{NG2021100041,ail_design_consideration}, which may in turn empower them to better recognize AI's persuasive intent \citep{carolus2023mailsmetaai,gallegos2025labelingmessagesaigenerateddoes,nature_ai_disinformation}. Going beyond existing AI literacy interventions that rely on static formats \citep{10.1145/3706598.3713254}, we propose \textbf{LLMimic}, a novel interactive tutorial in which users role-play as an LLM, walking through key stages of LLM training (pretraining, SFT, and RLHF). By having users mimic the LLM training process from a first-person perspective, LLMimic helps participants develop a deeper understanding of how LLMs generate outputs and how such outputs can become persuasive, manipulative, or biased. Its interactive, gamified design further enhances engagement. Moreover, while current AI literacy tools are typically evaluated only on self-reported AI literacy levels \citep{10.1145/3706598.3713254}, we go one step further and evaluate their downstream effect on behavior change.

We conducted a pre-registered\footnote{OSF preregistration: \url{https://osf.io/hb2mp/overview?view\_only=c7a15ac0aeec40a7bf19379837032c8c}} $2 \times 3$ between-subjects study to evaluate its effects on AI literacy and resistance to AI persuasion. The treatment group interacted with LLMimic, while the control group watched a video on AI history; participants in both conditions were then randomly assigned to one of three realistic persuasion tasks (charity donation, malicious money solicitation, and hotel recommendation). Through this study, we address the following research questions (RQs):

\textbf{RQ1}. Does exposure to LLMimic affect humans’ AI literacy and trust in AI?

\textbf{RQ2}. Does LLMimic mitigate the effects of persuasive AI?

\textbf{RQ3}. Do AI literacy and trust in AI mediate the relationship between exposure to LLMimic and persuasion outcomes?

The results show that LLMimic significantly improved participants' AI literacy, mitigated the effects of persuasive AI, and increased the truthfulness and social responsibility in ethical persuasion scenarios. In summary, we contribute: (1) \textbf{LLMimic}, a novel, interactive AI literacy tool that enables users to role-play key stages of LLM training; (2) empirical evidence that such an intervention mitigates the effects of persuasive AI in realistic human–AI interactions; and (3) design implications and practical guidelines for developing tools to enhance AI literacy and interventions to empower users to critically engage with increasingly persuasive AI systems.
\vspace{-4pt}


\section{Related Work}
\textbf{AI persuasion and mitigation.} LLMs can effectively persuade people in positive contexts (e.g., reducing conspiracy beliefs and political polarization) to promote social good \citep{llm_persuade_pol_issues,10.1145/3579592,coversational_persu,doi:10.1126/science.adq2852,wang2020persuasiongoodpersonalizedpersuasive,doi:10.1126/science.adq1814}. However, these capabilities also introduce risks, as LLMs may produce biased or inaccurate outputs that mislead users’ judgments and decisions \citep{doi:10.1073/pnas.2322823121,biased_AI_write_assi,potter-etal-2024-hidden}. Prior work has attempted to mitigate these risks through technical approaches, such as building classifiers to detect persuasive or manipulative intent, but these efforts remain limited \citep{wang2024mentalmanipdatasetfinegrainedanalysis,wilczyński2024resistancemanipulativeaikey,srivastav2026unknownunknownshiddenintentions,modzelewski2026aigeneratedpersuasiondetectedpersuaficial,kong-etal-2025-safepersuasion}. Human-centered approaches have also been explored, yet interventions such as labeling messages as AI-generated do not reliably reduce their influence \citep{gallegos2025labelingmessagesaigenerateddoes,biased_AI_write_assi}. These limitations motivate alternative human-centered strategies, with AI literacy emerging as a promising direction, similar to media literacy in misinformation research \citep{NG2021100041,ail_design_consideration,fakenewsgame,wang2024mentalmanipdatasetfinegrainedanalysis,wilczyński2024resistancemanipulativeaikey,fisher2024biasedaiinfluencepolitical}.

\textbf{AI literacy intervention.} Existing AI literacy efforts often rely on static materials to improve user discernment \citep{MARKUS2024100176, 10.1145/3706598.3713254, mlliteracy_reliance, 10.1145/3372782.3406252, KAJIWARA2024100251}. However, experiential approaches, such as role-playing, may foster deeper cognitive shifts and reduce automation bias in persuasive settings \citep{KAJIWARA2024100251, skitka1999does}. By exposing how AI generates content, these interventions could calibrate trust and encourage critical evaluation of communicative intent \citep{laupichler2025algorithm, wilczyński2024resistancemanipulativeaikey}. While game-based ``prebunking'' helps users identify manipulative techniques \citep{fakenewsgame, basol2020good, basol2021towards, roozenbeek2020breaking}, these methods focus on content cues rather than the AI source itself. Although structured engagement has been shown to support the evaluation of LLM behavior \citep{audit_llm, aici}, it remains unclear whether such interventions can mitigate the persuasive effects of AI. We present one of the first empirical studies of a literacy-centered intervention for AI persuasion.
\vspace{-6pt}

\section{Methodology}
\subsection{LLMimic: A role-play-based, interactive, gamified AI literacy tutorial}

\begin{wrapfigure}[23]{r}{0.5\linewidth}
\vspace{-12pt}
  \centering
  \includegraphics[width=\linewidth]{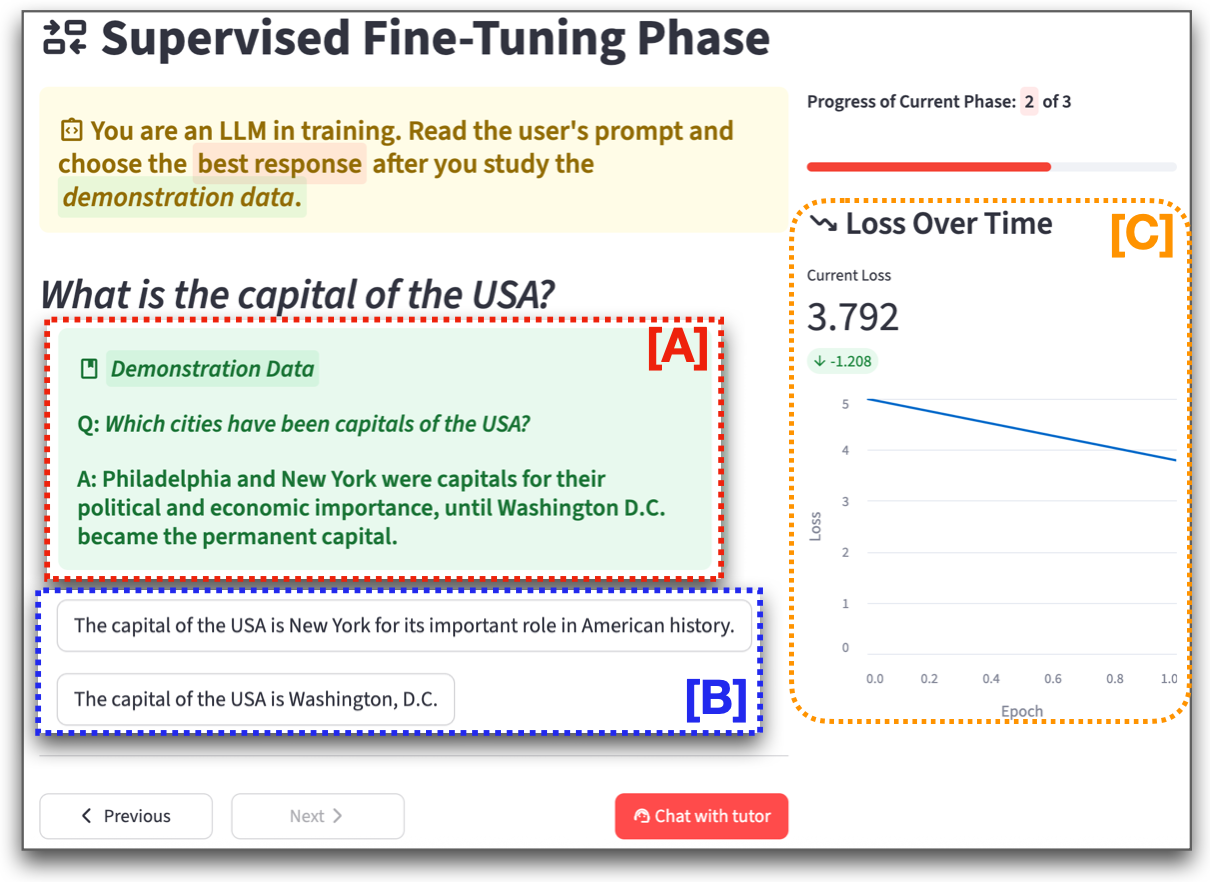}
  \caption{The LLMimic interface example. \fan{[A]} \textbf{Role-play-based}: Participants role-play as an LLM, progressing through training stages. \gee{[B]} \textbf{Interactive}: Participants answer questions and receive timely summary of key concepts. \orange{[C]} \textbf{Gamified}: As an LLM in training, participants observe real-time changes in their  \textit{loss} or \textit{reward}.}
  \label{fig:llmimic_demo}
\end{wrapfigure}

Following prior design considerations for AI literacy tutorials \citep{ail_design_consideration,NG2021100041}, we designed LLMimic to be (1) role-play-based, (2) gamified, and (3) interactive.

\textbf{(1) Role-play-based.} In LLMimic, participants role-play as an LLM and progress through the LLM training pipeline across three stages with curated, realistic examples: \textbf{Pre-training}, \textbf{Supervised Fine-tuning (SFT)}, and \textbf{Reinforcement Learning from Human Feedback (RLHF)} \citep{ouyang2022traininglanguagemodelsfollow}. 
For instance, in \textbf{Pre-training}, they perform token prediction with curated probabilities. In \textbf{SFT}, they answer Q\&A-style prompts by imitating \textit{Demonstration Data} (Figure~\ref{fig:llmimic_demo}\fan{[A]}). In \textbf{RLHF}, they compare two acceptable responses under a \textit{Reward Model}, and select the better one. See Appendix~\ref{appendix:llmimic_details} for details. 
This role-play-based design provides an immersive experience that enables users to better understand how LLMs generate content and why outputs may be biased or persuasive, without requiring technical backgrounds.

\textbf{(2) Interactive.} Before each stage, we provide an overview of its training purpose (e.g., instruction following for SFT). For each question, participants select the best output from multiple candidates (Figure~\ref{fig:llmimic_demo}\gee{[B]}) and then receive a brief \textit{Takeaway} that reinforces the key concepts (e.g., what is AI hallucination). We also provide an AI tutor that participants can interact with for follow-up questions. This interactive design ensures that participants receive reinforced and timely feedback.

\textbf{(3) Gamified.} As an LLM in training, participants also receive real-time changes in their \textit{loss} or \textit{reward} after selecting an answer (Figure~\ref{fig:llmimic_demo}\orange{[C]}). Selecting a wrong answer causes their \textit{loss} to increase, and vice versa. As they progress, the system highlights the capabilities they have acquired along the way (e.g., generating content, following instructions, answering questions, and producing human-aligned and safety-aware responses). This gamified design keeps the experience engaging.

\textbf{Persuasion-related content.} Besides standard training knowledge, we also tailored content related to persuasion and manipulation. In pre-training, we included examples of gender stereotypes (e.g., nurse as female) and manipulative examples from \cite{wang2024mentalmanipdatasetfinegrainedanalysis}. In SFT, participants \textit{generated} persuasive messages related to losing weight using credibility and logical appeals. In RLHF, LLMimic demonstrated how LLMs use emotional appeal and personalization in persuasion.
\vspace{-6pt}


\subsection{Persuasion scenarios}
We assessed LLMimic's effect on AI-driven persuasion using three scenarios spanning two dimensions: if the persuasion is active, and if the persuasion is ethical. \textit{Active persuasion} involves direct and explicit persuasive attempts to influence users’ actions (e.g., money solicitation), whereas \textit{passive persuasion} arises from how options are selected and presented (e.g., recommendations, ads) to subtly influence users, a common case in daily situations. We developed three persuasive agents for different persuasion tasks with \texttt{ChatGPT-4o}, and prompted agents to personalize their response based on participants' demographics.  Participants were randomly assigned to one of three persuasion scenarios:

(a) \textbf{Charity Donation} (denoted as \textit{Donation}; \textbf{active}, \textbf{ethical}), adapted from a previous study \citep{wang2020persuasiongoodpersonalizedpersuasive} as a classic example of ethical persuasion scenarios where the agent actively persuades the participant to donate to the charity \textit{Save the Children}. After the interaction, the participant will decide whether and how much to donate.

(b) \textbf{Malicious Money Solicitation} (denoted as \textit{MakeMePay}; \textbf{active}, \textbf{malicious}) was originally introduced by OpenAI to evaluate manipulative LLM behavior \citep{gpt4-5syscard}. In this scenario, the agent will solicit money by all means without a specific cause. We adopted this task to mimic potential fraudulent scenarios in the wild.

(c) \textbf{Hotel Booking} (denoted as \textit{Hotel}; \textbf{passive}, \textbf{ethical}) simulates an AI-powered booking system. Participants selected a three-night stay in Midtown New York City under a \$200/night budget, while the agent recommended hotels during the interaction. Passive persuasion was implemented by prioritizing certain hotels as \textit{``Featured''}. See Appendix~\ref{appendix:ui_demo} for details.

In the \textit{Donation} and \textit{MakeMePay} scenarios, participants interacted with the agent for six to ten turns, then decided whether to make a payment and, if so, specified an amount between \$0.01 and \$100.00. In the \textit{Hotel} scenario, participants selected one hotel from five options without a minimum turn requirement, reflecting typical real-world usage.
\vspace{-6pt}

\subsection{Data collection procedure} 
\begin{figure*}[ht]
  \centering
  \includegraphics[width=1\linewidth]{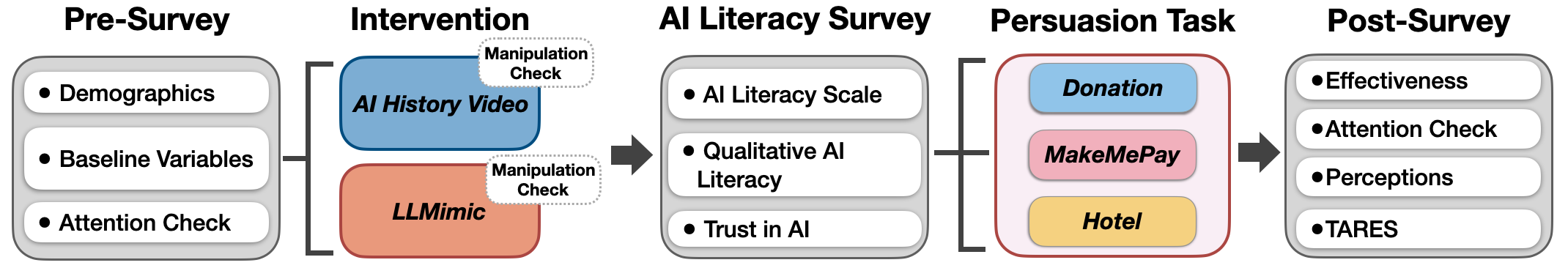}
  \vspace{-16pt}
  \caption{Human study flowchart. Participants completed a pre-survey, were randomly assigned to LLMimic or the control tutorial, then completed an AI literacy survey, one of three persuasion tasks, and a post-survey.
  }
  \label{fig:study_flow}
      \vspace{-14pt}
\end{figure*}

We conducted a $2 \times 3$ human study consisting of five stages (Figure~\ref{fig:study_flow}). (1) Participants completed a \textit{pre-survey} where we collected demographics and baseline measures related to AI use and persuasion (e.g., \textit{how often do you use generative AI?}). (2) In the \textit{intervention stage}, participants were randomly assigned to either the control or treatment condition. The control group watched a video tutorial on the history of AI\footnote{``A Short History of AI (11 minutes),'' Digital Management, University of Hohenheim, 2020. \url{https://www.youtube.com/watch?v=6QNOQrenue8}}, consistent with prior work \citep{10.1145/3706598.3713254}, while the treatment group interacted with LLMimic. Both groups completed the same manipulation check, confirming that LLMimic functioned as intended ($\chi^2(1)=7.54$, $p<.01$). (3) Participants then completed an \textit{AI literacy} survey, reporting their literacy, and optionally reflecting appropriate AI usage \citep{ail_design_consideration,PINSKI2024100062}. Next, participants were randomly assigned to (4) one of the three \textit{persuasion tasks}. Finally, participants completed (5) a \textit{post-survey} evaluating perceived agent quality. This design enabled sufficient data to address our research questions.

\subsection{Measures}

\label{sec:baseline_controls}
\textbf{Controlled variables.}
We controlled for AI- and persuasion-related factors, including \textit{AI experience}, \textit{AI expertise}, \textit{AI trust}, \textit{persuasion experience}, \textit{persuasion knowledge}, and\textit{ education level} in subsequent analyses. Before analysis, we confirmed that the control and treatment groups were comparable on these measures (see Appendix~\ref{appendix:measures_baseline}).

\textbf{AI literacy scale.} 
To assess AI literacy, we used a shortened 10-item version of the Meta AI Literacy Scale (MAILS) \citep{carolus2023mailsmetaai}, covering core competencies, AI ethics, self-efficacy, persuasion, and data literacy \citep{ail_design_consideration} (7-point Likert, see Appendix~\ref{appendix:mails_word} for details).

\textbf{Qualitative reflections on AI usage.} We included an optional open-ended question in the AI literacy survey asking participants when they would or would not use AI (see Appendix~\ref{appendix:mails_word} for details).

\textbf{Trust in AI.} We measured participants’ trust in AI twice using a 7-point Likert scale, (1) before the intervention and (2) after the intervention but prior to the persuasion tasks. 

\textbf{Persuasion outcome.}
\label{sec:rationale}
We defined a binary persuasion outcome based on whether a payment was made (Donation, MakeMePay) or the target hotel was selected (Hotel). We also recorded payment amounts, chosen hotels, and decision rationales. 

\textbf{TARES ethical persuasion score and perceptions of the agent.} We adopted the TARES principles of ethical persuasion to assess perceived ethical qualities: (1) \textbf{T}ruthfulness, (2) \textbf{A}uthenticity, (3) \textbf{R}espect, (4) \textbf{E}quity, and (5) \textbf{S}ocial responsibility \citep{tares}. 
We computed a composite TARES score by averaging the five items. In addition, we measured participants' perceptions of the agent across four dimensions: \textit{engagement, persuasiveness, perceived user autonomy}, and \textit{role fulfillment}. Item wordings are in Appendix~\ref{appendix:TARES_word}.
\vspace{-6pt}

\section{Results}
We recruited 313 adult, English-speaking participants residing in the United
States from Prolific, a crowdsourcing platform, between Dec. 10–11, 2025. Participants were compensated at \$12.00/hour. After excluding participants who failed attention checks or exceeded interaction limits, a final sample of $N = 274$ was retained for the data analysis. The sample comprised 51.09\% female and 47.45\% male participants, spanned ages 18–65+, and was 71.90\% White, 13.14\% Black, 4.74\% Mixed, and 2.92\% Asian (see Appendix~\ref{appendix:participants} for details).

\subsection{RQ1: LLMimic improves participants' AI literacy}
\begin{figure}[!ht]
  \centering
  \includegraphics[width=1\linewidth]{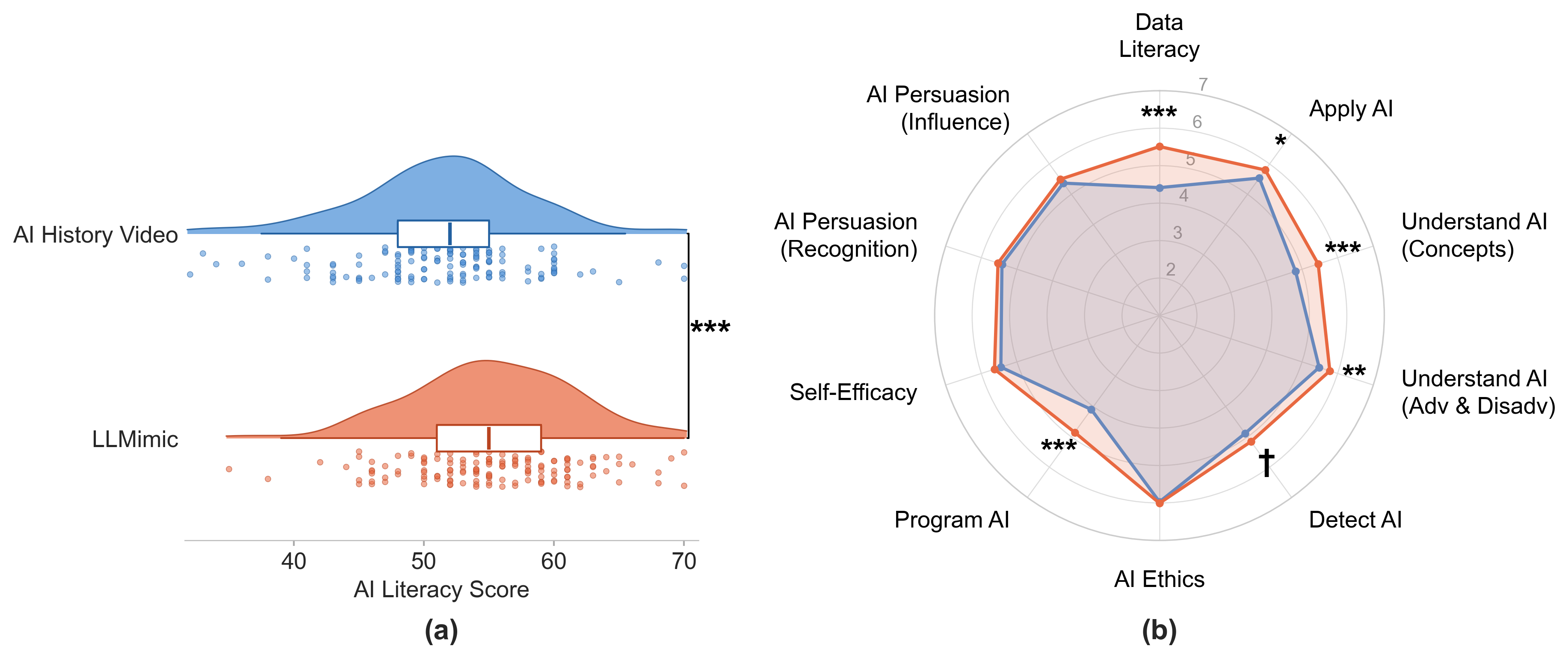}
    \vspace{-16pt}
    \caption{\textbf{(a)} The treatment group reported higher AI literacy than the control group. \textbf{(b)} At the item level, LLMimic improved \textit{Data Literacy}, \textit{Apply AI}, \textit{Understand AI}, and \textit{Program AI (select a useful tool to program an AI)}.$^{*}p<.05$, $^{**}p<.01$, $^{***}p<.001$, $.05<p^{\dagger}<.10$.}
  \label{fig:ail_res}
  \vspace{-10pt}
\end{figure}

\textbf{AI literacy and trust.}
Figure~\ref{fig:ail_res} shows that participants who interacted with LLMimic reported significantly higher AI literacy than those in the control condition (Treatment: $N = 141$, $M = 55.22$, $SD = 6.26$; Control: $N = 133$, $M = 51.44$, $SD = 6.76$), $p < .001$, and we also observed item-level improvements on AI literacy across multiple dimensions. On the trust side, participants in both groups experienced a significant decrease in trust after the intervention: from 4.91 to  4.54 for the control group ($p < .001$) and 5.13 to 4.70 for the treatment group ($p < .001$). The difference between the two groups was not statistically significant. This may stem from the fact that both interventions discuss the limitations of AI.


\textbf{Qualitative reflections on AI usage.}
Participants also shared optional qualitative reflections on when they would or would not use AI. Two researchers annotated participants' optional open-ended qualitative reflections into seven subcategories derived from common human-AI activities \citep{NIST_AI_TAX} (inter-annotator agreement score $>.7$). Results show that, compared to the control group, the treatment group more frequently identified content generation (53.3\% vs. 49.5\%) and personalization (7.8\% vs. 3.0\%) as appropriate AI use cases. Besides, they were less likely to view information discovery as appropriate (44.4\% vs. 50.5\%) and more likely to describe it as inappropriate (27.8\% vs. 23.2\%). This may suggest a more critical view of AI's role in information-related tasks. See Appendix~\ref{appendix:ail_qual} for more details. Moreover, participants also reported positive experiences with LLMimic and described it as “\textit{clear, engaging, and easy to understand,}” and noted that it \textit{"allowed [them] to understand [their] part in helping train these AI agents."}
\vspace{-10pt}

\subsection{RQ2: LLMimic mitigates the effects of persuasive AI}
\begin{figure}[h!]
\centering
\includegraphics[width=1\linewidth]{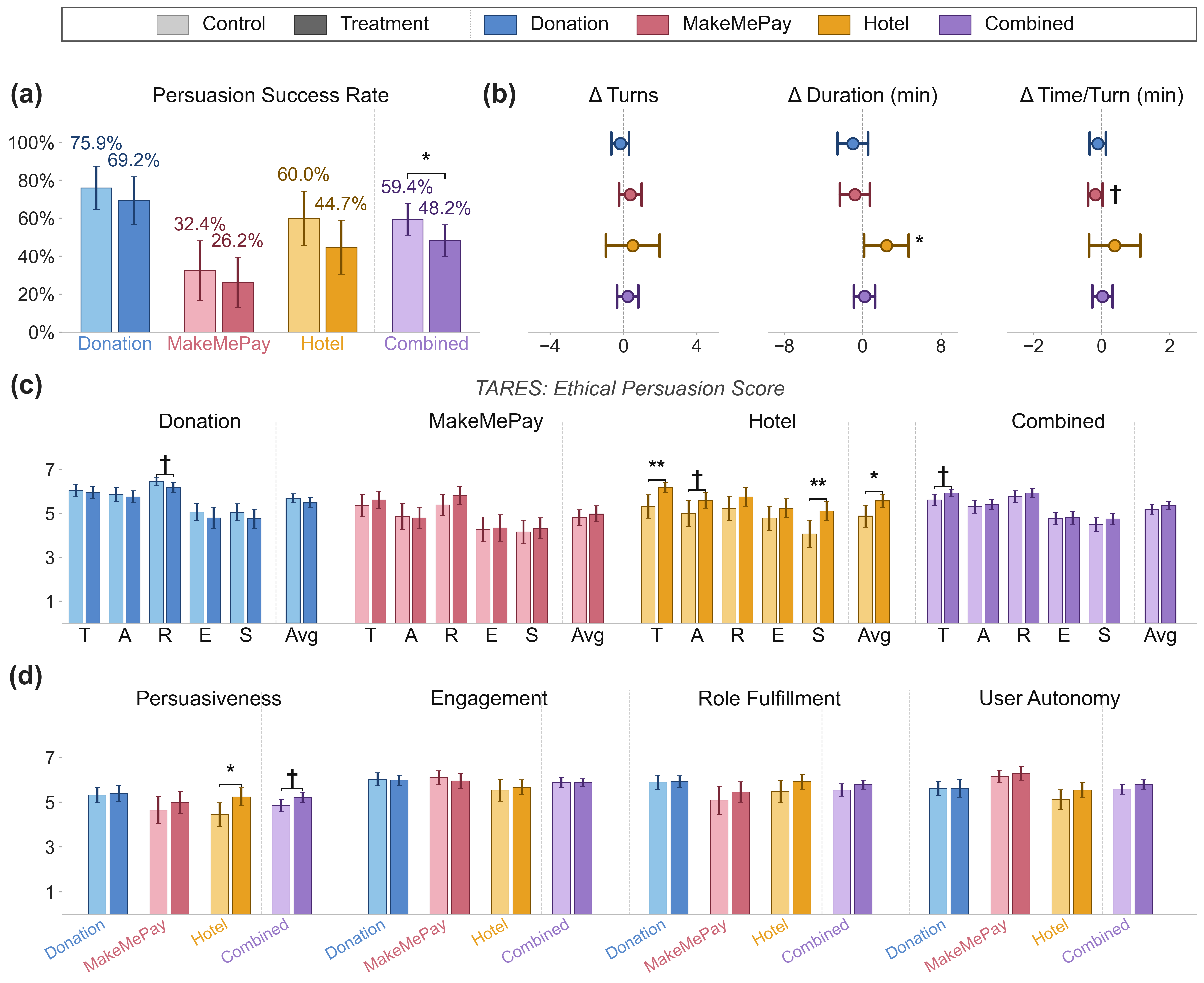}
\vspace{-16pt}
\caption{\textbf{(a)} Persuasion success rate across three scenarios and combined. The treatment group shows lower success rates across all scenarios. \textbf{(b)} Differences (Treatment $-$ Control) in persuasion interaction turns, duration, and average time per turn. Points indicate mean differences with 95\% CIs. \textbf{(c)} TARES ethical perception scores (\textbf{T}ruthfulness, \textbf{A}uthenticity, \textbf{R}espect, \textbf{E}quity, \textbf{S}ociety), and composite average score by scenario and condition. \textbf{(d)} Other perception ratings: Persuasiveness, Engagement, Role Fulfillment, and User Autonomy.}
\label{fig:results}
\end{figure}

\textbf{Effect on persuasion results.} Controlling for baseline variables (Section~\ref{sec:baseline_controls}), we fit a logistic regression model, and the results revealed that LLMimic significantly reduced overall persuasion success ($\beta = -0.54$, $OR = 0.58$, $p = .045$, Table~\ref{tab:logistic_combined}) across scenarios. On average, treatment participants exhibited 42\% lower odds of being persuaded compared to the control group, with descriptive reductions in persuasion observed across all scenarios (Figure~\ref{fig:results}(a)). These findings answer \textbf{RQ2}: LLMimic can help mitigate the effects of persuasive AI.

\begin{table}[!h]
\centering
\small
\setlength{\tabcolsep}{4pt}
\renewcommand{\arraystretch}{0.95}
\resizebox{\linewidth}{!}{%
\begin{tabular}{lcccccccc}
\toprule
& \multicolumn{4}{c}{\textbf{Model 1: w/o Scenario Control}}
& \multicolumn{4}{c}{\textbf{Model 2: w/ Scenario Control}} \\
\cmidrule(lr){2-5} \cmidrule(lr){6-9}
Predictor
& $b$ & SE & OR & 95\% CI
& $b$ & SE & OR & 95\% CI \\
\midrule
\textit{(Intercept)}
& $-$0.49 & 0.77 & 0.61 & [0.13, 2.78]
& 0.39    & 0.85 & 1.47 & [0.28, 7.83] \\
\textbf{Treatment}
& $\mathbf{-0.54^{*}}$ & \textbf{0.25} & \textbf{0.58} & \textbf{[0.35, 0.96]}
& $\mathbf{-0.54^{*}}$ & \textbf{0.27} & \textbf{0.58} & \textbf{[0.34, 0.99]} \\
AI Experience
& $-$0.02 & 0.12 & 0.98 & [0.77, 1.25]
& $-$0.06 & 0.13 & 0.95 & [0.73, 1.22] \\
AI Expertise
& $-$0.11 & 0.15 & 0.90 & [0.67, 1.20]
& $-$0.17 & 0.16 & 0.85 & [0.61, 1.16] \\
Persuasion Experience
& $0.23^{**}$  & 0.08 & 1.26 & [1.08, 1.47]
& $0.20^{*}$   & 0.08 & 1.22 & [1.04, 1.44] \\
Persuasion Strategy
& $-$0.21 & 0.26 & 0.81 & [0.49, 1.34]
& $-$0.23 & 0.28 & 0.80 & [0.46, 1.37] \\
AI Trust (pre)
& 0.11 & 0.11 & 1.12 & [0.90, 1.39]
& 0.16 & 0.12 & 1.18 & [0.94, 1.48] \\
Education
& 0.02 & 0.10 & 1.02 & [0.83, 1.25]
& 0.08 & 0.11 & 1.09 & [0.88, 1.36] \\
\midrule
Scenario: Hotel
& \multicolumn{4}{c}{---}
& $-0.87^{**}$  & 0.31 & 0.42 & [0.22, 0.76] \\
Scenario: MakeMePay
& \multicolumn{4}{c}{---}
& $-1.91^{***}$ & 0.35 & 0.15 & [0.07, 0.29] \\
\midrule
AIC ($\downarrow$)
& \multicolumn{4}{c}{377.80}
& \multicolumn{4}{c}{348.53} \\
\bottomrule
\end{tabular}%
}
\caption{
LLMimic significantly reduced overall persuasion success. Model 1: $\text{Persuasion} = \beta_0 + \beta_1\text{Treatment} + \boldsymbol{\beta}_2\mathbf{X}$, where $\mathbf{X}$ denotes baseline controls. Model 2 additionally controls for scenario variables (Donation as reference). AIC ($\downarrow$): Akaike Information Criterion.
}
\label{tab:logistic_combined}
\end{table}

Interaction patterns also differed between conditions (Figure~\ref{fig:results}b). In the Hotel scenario, they spent significantly more time interacting with the agent (Treatment: $M = 9.46$; Control: $M = 7.04$; $p = .039$). In the MakeMePay scenario, they tended to spend less time per round ($M = 0.98$ vs. $1.16$; $p = .1$) but engaged in slightly more total turns ($M = 7.31$ vs. $6.94$; $p = .242$). Furthermore, among participants who were persuaded by agents, the treatment group donated more in the ethical Donation scenario ($M = 71.81$ vs. $62.54$) but paid less in the malicious MakeMePay scenario ($M = 43.18$ vs. $60.82$). This pattern suggests that participants responded differently to ethical versus potentially malicious requests, rather than uniformly reducing spending.

\textbf{Effect on perception.}
Figure~\ref{fig:results}(c) shows the perceived TARES ethical persuasion score across scenarios: Donation was perceived more ethical than MakeMePay ($p < .001$), while differences between Donation and Hotel ($p = .07$) and between Hotel and MakeMePay ($p = .07$) were marginal. Within the Hotel scenario, treatment participants perceived the agent as more ethical than those in the control condition (Control: $M = 4.88$, $SD = 1.73$; Treatment: $M = 5.57$, $SD = 1.08$; $p = .02$). This pattern was also reflected at the dimension level, with higher ratings in \textit{truthfulness} ($M = 6.17$ vs.\ $5.31$, $p = .005$) and \textit{social responsibility} ($M = 5.11$ vs.\ $4.07$, $p = .008$). These findings suggest that although our LLMimic intervention reduces persuasion success, it also leads the persuasion to be perceived as more truthful and socially responsible in the ethical scenario of hotel recommendation. Besides,  passive persuasion was perceived as more acceptable, potentially due to more structured and objective information presented (e.g., hotel cards). 
For instance, participants describe the system as “\textit{clear and helpful...,}” and suggest that it would be “\textit{even more useful}” with side-by-side comparisons. This points to the value of more structured and transparent information presentation in recommendation settings, which may better support users’ agency.

On other perceptions, participants in the treatment group tended to perceive AI agents as more persuasive (Control: $M = 4.85$, $SD = 1.79$; Treatment: $M = 5.21$, $SD = 1.40$; $p=.052$). Notably, the Hotel agent was rated as significantly more persuasive by participants in the treatment group (Control: $M = 4.44$, $SD = 1.79$; Treatment: $M = 5.23$, $SD = 1.40$; $p = .02$). We also observed descriptive differences for \textit{role fulfillment} and \textit{user autonomy}, with both dimensions rated descriptively higher in the treatment group (see Figure~\ref{fig:results}(d)). Overall, both TARES and other post-experiment perception measures (except for \textit{Equity}) were significantly associated with persuasion decisions (see Appendix~\ref{appendix:direct_tares} and~\ref{appendix:direct_perc} for more detail). 

\textbf{Persuasion strategy.} We used \texttt{ChatGPT-4o} to annotate sentence-level persuasion strategies based on a prior taxonomy \citep{zeng2024johnnypersuadellmsjailbreak}, adding a non-strategy category (e.g., greetings and chit-chat; accuracy = .88 based on manual verification of $n = 210$ instances by one author). Excluding the non-strategy category, different scenarios employed different persuasion strategies. The Donation agent relied on \textit{information-} (10.2\%), \textit{credibility-} (8.9\%), \textit{relationship-} (7.8\%), and \textit{emotion-based} (7.1\%) strategies. In contrast, the MakeMePay agent used \textit{information-\textbf{bias}} (9.4\%), \textit{relationship-} (8.9\%), and \textit{scarcity-based} (6.4\%) strategies. The Hotel agent relied on \textit{information-\textbf{bias}} (23.2\%), \textit{scarcity-} (15\%), and \textit{emotion-based} techniques (7.6\%); (Appendix~\ref{appendix:strategy_distribution}). These differences aligned user perceptions: the treatment group perceived the Donation agent as marginally more respectful ($p < .10$), the Hotel agent as more persuasive, and the MakeMePay agent as less ethical. See Appendix~\ref{appendix:tares_scenario_diff} for details.

\textbf{Persuasion decision rationale.} Participants also shared their rationales when they made decisions in the persuasion scenarios, and two researchers independently annotated the rationales into three categories: \textit{practical} (e.g., hotel rating), \textit{personal} (e.g., \textit{“I want to help”}), and \textit{AI-related} (e.g., \textit{“AI recommended...”}), with an inter-annotator agreement of 0.83.  Across scenarios, participants mentioned non-AI-related reasons more than AI-related ones. In the ethical Donation and Hotel contexts, both the acceptance and rejection of persuasion were largely driven by personal (e.g., ``care about children'') or practical considerations (e.g., ``within budget''), with fewer AI-related rationales. Descriptively, however, participants in the treatment group more frequently cited AI-related rationales when accepting persuasion in ethical contexts. 
In the unethical MakeMePay scenario, treatment participants more often cited AI-related reasons when rejecting the agent (e.g., ``the agent kept repeating''), which may indicate more explicit attention to AI-related cues in this malicious context. See Table~\ref{tab:rationale_examples} for example rationales and Appendix~\ref{appendix:rationale_factors} for full details.
\vspace{-8pt}
\begin{table}[!ht]
\centering
\scriptsize
\setlength{\tabcolsep}{2pt}
\renewcommand{\arraystretch}{0.95}
\begin{tabularx}{\linewidth}{p{1cm}p{1.05cm}p{1.3cm}p{0.9cm}X}
\toprule
\textbf{Category} & \textbf{Condition} & \textbf{Scenario} & \textbf{Decision} & \textbf{Rationale} \\
\midrule
\multirow{2}{*}{Practical}
& Control & Hotel & Accept & \textit{``It has all the amenities I am looking for and is \textcolor{purple}{within my budget}.''} \\
\cmidrule(lr){2-5}
& Control & Hotel & Reject & \textit{``\textcolor{purple}{Great location and fits the budget.}''} \\
\midrule
\multirow{2}{*}{Personal}
& Treatment & Donation & Accept & \textit{``I \textcolor{purple}{care about} children not suffering. I know what it’s like to struggle and be without.''} \\
\cmidrule(lr){2-5}
& Control & Donation & Reject & \textit{``I \textcolor{purple}{feel like} making a bigger donation would not be a good idea since the Ceo and staff are making way too much money''} \\
\midrule
\multirow{4}{*}{AI}
& Treatment & Donation & Accept & \textit{``...\textcolor{purple}{gave me enough information} ... I feel comfortable donating to the cause.''} \\
\cmidrule(lr){2-5}
& Treatment & MakeMePay & Accept & \textit{``The \textcolor{purple}{model says} I can be refunded if it doesn’t work out, so I could recover my losses.''} \\
\cmidrule(lr){2-5}
& Treatment & MakeMePay & Reject & \textit{``Jessie’s \textcolor{purple}{responses were vague}... low-content discourse was unconvincing and off-putting.''} \\
\bottomrule
\end{tabularx}
\caption{Examples of participant decision rationales across conditions and scenarios}
\label{tab:rationale_examples}
\vspace{-10pt}
\end{table}

\subsection{RQ3: Exploring the underlying mechanisms of LLMimic on persuasion}
Our mediation analysis with structural equation modeling (SEM) shows that 
AI literacy and trust did not mediate the intervention effect. The intervention significantly improved AI literacy ($\beta = .27$, $p < .001$) but not AI trust ($\beta = -.01$, $p = .86$), and neither AI literacy nor trust predicted persuasion outcomes (AI literacy: $\beta = -.04$, $p = .72$; trust: $\beta = .04$, $p = .62$). No indirect effects were observed (AI literacy: $\beta = -.01$, $p = .72$; trust: $\beta = -.00$, $p = .86$), providing no support for \textbf{RQ3}, see Appendix~\ref{appendix:mediation} for details. This suggests that the reduction in persuasion is not captured by AI literacy and trust scales measured in this study, and we discuss the potential mechanisms in the following discussion section.
\vspace{-4pt}

\section{Discussion}
\textbf{Key Findings.} Before discussing the broader implications, we summarize two key findings:
\begin{itemize}[leftmargin=*,itemsep=1pt,topsep=2pt]
    \item \textbf{LLMimic significantly improved participants' AI literacy} and \textbf{reduced overall susceptibility to persuasive AI across realistic scenarios}, although neither self-reported AI literacy nor AI trust mediated this effect.
    \item \textbf{Participants did not simply reject AI persuasion;} instead, their responses showed signs of greater discernment, including resistance to potentially manipulative persuasion while perceiving ethical AI recommendations more positively.
\end{itemize}
\textbf{LLMimic mitigates AI persuasion through mechanisms beyond self-reported AI literacy and trust.}
Our findings show that LLMimic significantly improved participants' AI literacy and reduced their overall odds of being persuaded by AI. While we examined whether self-reported AI literacy and trust mediated this effect, neither pathway was supported by the mediation analysis. 
This suggests that the observed reduction in persuasion is not captured by self-reported AI literacy and trust \textit{\textbf{scales}}, but may instead reflect more fine-grained shifts in how participants respond to persuasive AI. This interpretation is broadly consistent with our qualitative and perception results, including higher perceived persuasiveness in the passive context and more frequent AI-related rationales in the malicious persuasion scenario. These findings point to the need for future work to investigate how LLMimic builds resistance to persuasive AI and shapes people’s judgments and perceptions. Despite this, our results suggest that even a brief intervention ($\sim$15 minutes) can improve AI literacy and mitigate the effects of AI persuasion across the tested contexts. Future work should also include longitudinal studies to assess its durability over time. More broadly, these findings demonstrate the value of strengthening public AI literacy education as part of broader efforts to protect people from malicious AI persuasion.


\textbf{Mitigating harms from AI persuasion involves both resistance and discernment.}
Our results show that LLMimic reduced persuasion success across both ethical and unethical scenarios, suggesting that the intervention builds a general resistance that does not distinguish between manipulative and legitimate persuasive intent. We acknowledge this as a potential limitation, as AI persuasion can also support positive outcomes (e.g., encouraging physical exercise \citep{liang2021evaluation, oh2025enhancing}) and LLMimic may reduce such positive effects. That said, our findings offer encouraging signals: participants perceived the ethical hotel recommendation as more truthful and socially responsible (as reflected in TARES scores), indicating that the distinction between ethical and unethical persuasion is not entirely lost. To develop more targeted mitigation that preserves beneficial uses of AI persuasion, future work should foster not only resistance but also discernment, for instance, exploring how to empower users to differentiate between malicious and prosocial AI persuasion.

\textbf{AI-AI evaluations do not directly reflect human susceptibility to AI persuasion.}
Our study suggests that AI-AI evaluations may overestimate human susceptibility to AI persuasion in the MakeMePay scenario \citep{gpt4-5syscard}. OpenAI reports up to 57\% persuasion success under this setting on \texttt{ChatGPT-4o}, whereas only 32.4\% of participants in our control group were persuaded. One possible reason is that human recipients are not passive targets: they can interpret persuasive intent, evaluate credibility, and respond according to their judgments. This interpretation is broadly consistent with our perception results (Appendix~\ref{appendix:direct_perc} and~\ref{appendix:direct_tares}), suggesting that persuasion outcomes are related to participants' subjective evaluations. Taken together, these findings suggest that AI-AI benchmark evaluations should be interpreted with caution as direct proxies for human susceptibility to persuasion. Our study also contributes a human-centered evaluation perspective by applying TARES to assess how participants perceived the ethical qualities of persuasive AI. Alongside the behavioral results, these perception measures highlight the value of complementing AI-AI benchmarks with human-centered evaluations in future research on persuasive AI and AI safety.
\vspace{-6pt}

\section{Conclusion}
We present LLMimic, an interactive, gamified AI literacy tutorial in which participants take the perspective of an LLM and progress through key training stages. Designed for non-technical users, LLMimic was evaluated in a human-subjects study. Results show that LLMimic improves AI literacy, reduces persuasion success across three realistic scenarios, and increases perceived truthfulness and social responsibility in the Hotel scenario. These findings suggest that such proactive, human-centered interventions can help mitigate potentially malicious AI persuasion and support more informed user interactions at scale. Future work should examine its longitudinal effects and focus on helping people discern between malicious and prosocial AI persuasion.


\section*{Ethics Statement}
\paragraph{IRB Review} This study was approved by the Institutional Review Board (IRB) at Northeastern University, USA (Protocol \#25-05-43). All participants provided informed consent prior to participation and were free to withdraw at any time. We explicitly informed participants that the persuasion scenarios were simulated and that no real money would be deducted, while instructing them to treat the tasks as if they were real. Participants were compensated at an hourly rate of \$12.00.

\paragraph{Limitations} Our survey design may introduce potential confounds, as participants reported prior AI and persuasion experience alongside AI literacy and trust measures. Future work could employ a more closely matched control condition to better disentangle the effects of intervention content and design. In addition, although prompts were held constant, differences in how participants interacted with the agent may have led to variation in persuasion power. Overall, these limitations call for cautious interpretation, but do not undermine the main findings on LLMimic’s effectiveness.

\paragraph{Acknowledgments} We thank the Schmidt Sciences AI2050 Program for its support. Any opinions, findings, conclusions, or recommendations expressed in this material are those of the authors and do not necessarily reflect the views of the funding agency.

\bibliographystyle{colm2026_conference}
\bibliography{references}

@misc{wang2020persuasiongoodpersonalizedpersuasive,
      title={Persuasion for Good: Towards a Personalized Persuasive Dialogue System for Social Good}, 
      author={Xuewei Wang and Weiyan Shi and Richard Kim and Yoojung Oh and Sijia Yang and Jingwen Zhang and Zhou Yu},
      year={2020},
      eprint={1906.06725},
      archivePrefix={arXiv},
      primaryClass={cs.CL},
      url={https://arxiv.org/abs/1906.06725}, 
}

@article{laupichler2025algorithm,
  title={Algorithm aversion revisited: The role of AI literacy and attitudes towards AI in shaping perceptions of AI-generated texts},
  author={Laupichler, Matthias Carl and Knoth, Nils and Schleiss, Johannes and Raupach, Tobias},
  journal={British Journal of Educational Technology},
  year={2025},
  publisher={Wiley Online Library}
}

@article{skitka1999does,
  title={Does automation bias decision-making?},
  author={Skitka, Linda J and Mosier, Kathleen L and Burdick, Mark},
  journal={International Journal of Human-Computer Studies},
  volume={51},
  number={5},
  pages={991--1006},
  year={1999},
  publisher={Elsevier}
}

@article{basol2021towards,
  title={Towards psychological herd immunity: Cross-cultural evidence for two prebunking interventions against COVID-19 misinformation},
  author={Basol, Melisa and Roozenbeek, Jon and Berriche, Manon and Uenal, Fatih and McClanahan, William P and Linden, Sander van der},
  journal={Big Data \& Society},
  volume={8},
  number={1},
  pages={20539517211013868},
  year={2021},
  publisher={SAGE Publications Sage UK: London, England}
}

@article{roozenbeek2020breaking,
  title={Breaking Harmony Square: A game that “inoculates” against political misinformation},
  author={Roozenbeek, Jon and van der Linden, Sander},
  year={2020},
  publisher={Shorenstein Center for Media, Politics and Public Policy, at Harvard~…}
}

@article{basol2020good,
  title={Good news about bad news: Gamified inoculation boosts confidence and cognitive immunity against fake news},
  author={Basol, Melisa and Roozenbeek, Jon and Van der Linden, Sander},
  journal={Journal of cognition},
  volume={3},
  number={1},
  pages={2},
  year={2020}
}

@article{doi:10.1126/science.adq2852,
    author = {Michael Henry Tessler  and Michiel A. Bakker  and Daniel Jarrett  and Hannah Sheahan  and Martin J. Chadwick  and Raphael Koster  and Georgina Evans  and Lucy Campbell-Gillingham  and Tantum Collins  and David C. Parkes  and Matthew Botvinick  and Christopher Summerfield },
    title = {AI can help humans find common ground in democratic deliberation},
    journal = {Science},
    volume = {386},
    number = {6719},
    pages = {eadq2852},
    year = {2024},
    doi = {10.1126/science.adq2852},
    URL = {https://www.science.org/doi/abs/10.1126/science.adq2852},
    eprint = {https://www.science.org/doi/pdf/10.1126/science.adq2852},
    }

@article{
doi:10.1126/science.adq1814,
author = {Thomas H. Costello  and Gordon Pennycook  and David G. Rand },
title = {Durably reducing conspiracy beliefs through dialogues with AI},
journal = {Science},
volume = {385},
number = {6714},
pages = {eadq1814},
year = {2024},
doi = {10.1126/science.adq1814},
URL = {https://www.science.org/doi/abs/10.1126/science.adq1814},
eprint = {https://www.science.org/doi/pdf/10.1126/science.adq1814}}

@article{NG2021100041,
	author = {Davy Tsz Kit Ng and Jac Ka Lok Leung and Samuel Kai Wah Chu and Maggie Shen Qiao},
	doi = {https://doi.org/10.1016/j.caeai.2021.100041},
	issn = {2666-920X},
	journal = {Computers and Education: Artificial Intelligence},
	pages = {100041},
	title = {Conceptualizing AI literacy: An exploratory review},
	url = {https://www.sciencedirect.com/science/article/pii/S2666920X21000357},
	volume = {2},
	year = {2021}}

@inproceedings{ail_design_consideration,
    author = {Long, Duri and Magerko, Brian},
    title = {What is AI Literacy? Competencies and Design Considerations},
    year = {2020},
    isbn = {9781450367080},
    publisher = {Association for Computing Machinery},
    address = {New York, NY, USA},
    url = {https://doi.org/10.1145/3313831.3376727},
    doi = {10.1145/3313831.3376727},
    booktitle = {Proceedings of the 2020 CHI Conference on Human Factors in Computing Systems},
    location = {Honolulu, HI, USA},
    series = {CHI '20}
}

@misc{fisher2024biasedaiinfluencepolitical,
      title={Biased AI can Influence Political Decision-Making}, 
      author={Jillian Fisher and Shangbin Feng and Robert Aron and Thomas Richardson and Yejin Choi and Daniel W. Fisher and Jennifer Pan and Yulia Tsvetkov and Katharina Reinecke},
      year={2024},
      eprint={2410.06415},
      archivePrefix={arXiv},
      primaryClass={cs.HC},
      url={https://arxiv.org/abs/2410.06415}, 
}

@article{doi:10.1073/pnas.2322823121,
	author = {Matthew R. DeVerna and Harry Yaojun Yan and Kai-Cheng Yang and Filippo Menczer},
	doi = {10.1073/pnas.2322823121},
	eprint = {https://www.pnas.org/doi/pdf/10.1073/pnas.2322823121},
	journal = {Proceedings of the National Academy of Sciences},
	number = {50},
	pages = {e2322823121},
	title = {Fact-checking information from large language models can decrease headline discernment},
	url = {https://www.pnas.org/doi/abs/10.1073/pnas.2322823121},
	volume = {121},
	year = {2024}}

@misc{wilczyński2024resistancemanipulativeaikey,
      title={Resistance Against Manipulative AI: key factors and possible actions}, 
      author={Piotr Wilczyński and Wiktoria Mieleszczenko-Kowszewicz and Przemysław Biecek},
      year={2024},
      eprint={2404.14230},
      archivePrefix={arXiv},
      primaryClass={cs.HC},
      url={https://arxiv.org/abs/2404.14230}, 
}

@misc{wang2024mentalmanipdatasetfinegrainedanalysis,
      title={MentalManip: A Dataset For Fine-grained Analysis of Mental Manipulation in Conversations}, 
      author={Yuxin Wang and Ivory Yang and Saeed Hassanpour and Soroush Vosoughi},
      year={2024},
      eprint={2405.16584},
      archivePrefix={arXiv},
      primaryClass={cs.CL},
      url={https://arxiv.org/abs/2405.16584}, 
}

@misc{carolus2023mailsmetaai,
      title={MAILS -- Meta AI Literacy Scale: Development and Testing of an AI Literacy Questionnaire Based on Well-Founded Competency Models and Psychological Change- and Meta-Competencies}, 
      author={Astrid Carolus and Martin Koch and Samantha Straka and Marc Erich Latoschik and Carolin Wienrich},
      year={2023},
      eprint={2302.09319},
      archivePrefix={arXiv},
      primaryClass={cs.AI},
      url={https://arxiv.org/abs/2302.09319}, 
}

@article{MARKUS2024100176,
	author = {Andr{\'e} Markus and Jan Pfister and Astrid Carolus and Andreas Hotho and Carolin Wienrich},
	doi = {https://doi.org/10.1016/j.caeo.2024.100176},
	issn = {2666-5573},
	journal = {Computers and Education Open},
	pages = {100176},
	title = {Effects of AI understanding-training on AI literacy, usage, self-determined interactions, and anthropomorphization with voice assistants},
	url = {https://www.sciencedirect.com/science/article/pii/S266655732400017X},
	volume = {6},
	year = {2024}}

@inproceedings{potter-etal-2024-hidden,
    title = "Hidden Persuaders: {LLM}s' Political Leaning and Their Influence on Voters",
    author = "Potter, Yujin  and
      Lai, Shiyang  and
      Kim, Junsol  and
      Evans, James  and
      Song, Dawn",
    editor = "Al-Onaizan, Yaser  and
      Bansal, Mohit  and
      Chen, Yun-Nung",
    booktitle = "Proceedings of the 2024 Conference on Empirical Methods in Natural Language Processing",
    month = nov,
    year = "2024",
    address = "Miami, Florida, USA",
    publisher = "Association for Computational Linguistics",
    url = "https://aclanthology.org/2024.emnlp-main.244/",
    doi = "10.18653/v1/2024.emnlp-main.244",
    pages = "4244--4275"
}

@article{10.1145/3579592,
author = {Karinshak, Elise and Liu, Sunny Xun and Park, Joon Sung and Hancock, Jeffrey T.},
title = {Working With AI to Persuade: Examining a Large Language Model's Ability to Generate Pro-Vaccination Messages},
year = {2023},
issue_date = {April 2023},
publisher = {Association for Computing Machinery},
address = {New York, NY, USA},
volume = {7},
number = {CSCW1},
url = {https://doi.org/10.1145/3579592},
doi = {10.1145/3579592},
journal = {Proc. ACM Hum.-Comput. Interact.},
month = apr,
articleno = {116},
numpages = {29}
}

@misc{gallegos2025labelingmessagesaigenerateddoes,
      title={Labeling Messages as AI-Generated Does Not Reduce Their Persuasive Effects}, 
      author={Isabel O. Gallegos and Chen Shani and Weiyan Shi and Federico Bianchi and Izzy Gainsburg and Dan Jurafsky and Robb Willer},
      year={2025},
      eprint={2504.09865},
      archivePrefix={arXiv},
      primaryClass={cs.CY},
      url={https://arxiv.org/abs/2504.09865}, 
}

@article{PINSKI2024100062,
	author = {Marc Pinski and Alexander Benlian},
	doi = {https://doi.org/10.1016/j.chbah.2024.100062},
	issn = {2949-8821},
	journal = {Computers in Human Behavior: Artificial Humans},
	number = {1},
	pages = {100062},
	title = {AI literacy for users -- A comprehensive review and future research directions of learning methods, components, and effects},
	url = {https://www.sciencedirect.com/science/article/pii/S2949882124000227},
	volume = {2},
	year = {2024}}

@inproceedings{10.1145/3706598.3713254,
author = {Cao, Huajie Jay and Lee, Hee Rin and Peng, Wei},
title = {Empowering Adults with AI Literacy: Using Short Videos to Transform Understanding and Harness Fear for Critical Thinking},
year = {2025},
isbn = {9798400713941},
publisher = {Association for Computing Machinery},
address = {New York, NY, USA},
url = {https://doi.org/10.1145/3706598.3713254},
doi = {10.1145/3706598.3713254},
booktitle = {Proceedings of the 2025 CHI Conference on Human Factors in Computing Systems},
articleno = {203},
numpages = {8},
location = {
},
series = {CHI '25}
}

@article{costello2026large,
  title={Large language models can effectively convince people to believe conspiracies},
  author={Costello, Thomas H and Pelrine, Kellin and Kowal, Matthew and Arechar, Antonio A and Godbout, Jean-Fran{\c{c}}ois and Gleave, Adam and Rand, David and Pennycook, Gordon},
  journal={arXiv preprint arXiv:2601.05050},
  year={2026}
}

@article{oh2025enhancing,
  title={Enhancing physical activity through a relational artificial intelligence chatbot: A feasibility and usability study},
  author={Oh, Yoo Jung and Liang, Kai-Hui and Kim, Diane Dagyong and Zhang, Xuanming and Yu, Zhou and Fukuoka, Yoshimi and Zhang, Jingwen},
  journal={Digital Health},
  volume={11},
  pages={20552076251324445},
  year={2025},
  publisher={SAGE Publications Sage UK: London, England}
}

@inproceedings{liang2021evaluation,
  title={Evaluation of in-person counseling strategies to develop physical activity chatbot for women},
  author={Liang, Kai-Hui and Lange, Patrick L and Oh, Yoo Jung and Zhang, Jingwen and Fukuoka, Yoshimi and Yu, Zhou},
  booktitle={Proceedings of the 22nd annual meeting of the special interest group on discourse and dialogue},
  pages={32--44},
  year={2021}
}

@inproceedings{mlliteracy_reliance,
author = {Chiang, Chun-Wei and Yin, Ming},
title = {Exploring the Effects of Machine Learning Literacy Interventions on Laypeople’s Reliance on Machine Learning Models},
year = {2022},
isbn = {9781450391443},
publisher = {Association for Computing Machinery},
address = {New York, NY, USA},
url = {https://doi.org/10.1145/3490099.3511121},
doi = {10.1145/3490099.3511121},
booktitle = {Proceedings of the 27th International Conference on Intelligent User Interfaces},
pages = {148–161},
numpages = {14},
location = {Helsinki, Finland},
series = {IUI '22}
}

@inproceedings{10.1145/3372782.3406252,
author = {Register, Yim and Ko, Amy J.},
title = {Learning Machine Learning with Personal Data Helps Stakeholders Ground Advocacy Arguments in Model Mechanics},
year = {2020},
isbn = {9781450370929},
publisher = {Association for Computing Machinery},
address = {New York, NY, USA},
url = {https://doi.org/10.1145/3372782.3406252},
doi = {10.1145/3372782.3406252},
booktitle = {Proceedings of the 2020 ACM Conference on International Computing Education Research},
pages = {67–78},
numpages = {12},
location = {Virtual Event, New Zealand},
series = {ICER '20}
}

@article{fakenewsgame,
	author = {Roozenbeek, Jon and van der Linden, Sander},
	date = {2019/06/25},
	doi = {10.1057/s41599-019-0279-9},
	id = {Roozenbeek2019},
	isbn = {2055-1045},
	journal = {Palgrave Communications},
	number = {1},
	pages = {65},
	title = {Fake news game confers psychological resistance against online misinformation},
	url = {https://doi.org/10.1057/s41599-019-0279-9},
	volume = {5},
	year = {2019}}

@article{KAJIWARA2024100251,
	author = {Yusuke Kajiwara and Kouhei Kawabata},
	doi = {https://doi.org/10.1016/j.caeai.2024.100251},
	issn = {2666-920X},
	journal = {Computers and Education: Artificial Intelligence},
	pages = {100251},
	title = {AI literacy for ethical use of chatbot: Will students accept AI ethics?},
	url = {https://www.sciencedirect.com/science/article/pii/S2666920X24000547},
	volume = {6},
	year = {2024}}

@misc{ouyang2022traininglanguagemodelsfollow,
      title={Training language models to follow instructions with human feedback}, 
      author={Long Ouyang and Jeff Wu and Xu Jiang and Diogo Almeida and Carroll L. Wainwright and Pamela Mishkin and Chong Zhang and Sandhini Agarwal and Katarina Slama and Alex Ray and John Schulman and Jacob Hilton and Fraser Kelton and Luke Miller and Maddie Simens and Amanda Askell and Peter Welinder and Paul Christiano and Jan Leike and Ryan Lowe},
      year={2022},
      eprint={2203.02155},
      archivePrefix={arXiv},
      primaryClass={cs.CL},
      url={https://arxiv.org/abs/2203.02155}, 
}

@misc{gpt4-5syscard,
	author = {OpenAI},
	title = {{O}pen{A}{I} {G}{P}{T}-4.5 {S}ystem {C}ard},
	howpublished = {\url{https://cdn.openai.com/gpt-4-5-system-card-2272025.pdf}},
	year = {2025},
}

@misc{tares,
	author = {Sherry Baker, Brigham Young University},
	title = {The TARES Test: Five Principles for Ethical Persuasion},
	howpublished ={\url{http://www.communicationcache.com/uploads/1/0/8/8/10887248/the_tares_test-_five_principles_for_ethical_persuasion.pdf}},
	year = {2021},
}

@article{aici,
	author = {Zhang, Helen and Perry, Anthony and Lee, Irene},
	date = {2025/03/01},
	doi = {10.1007/s40593-024-00398-x},
	id = {Zhang2025},
	isbn = {1560-4306},
	journal = {International Journal of Artificial Intelligence in Education},
	number = {1},
	pages = {398--438},
	title = {Developing and Validating the Artificial Intelligence Literacy Concept Inventory: an Instrument to Assess Artificial Intelligence Literacy among Middle School Students},
	url = {https://doi.org/10.1007/s40593-024-00398-x},
	volume = {35},
	year = {2025}}

@article{llm_persuade_pol_issues,
	author = {Bai, Hui and Voelkel, Jan G. and Muldowney, Shane and Eichstaedt, Johannes C. and Willer, Robb},
	date = {2025/07/01},
	doi = {10.1038/s41467-025-61345-5},
	id = {Bai2025},
	isbn = {2041-1723},
	journal = {Nature Communications},
	number = {1},
	pages = {6037},
	title = {LLM-generated messages can persuade humans on policy issues},
	url = {https://doi.org/10.1038/s41467-025-61345-5},
	volume = {16},
	year = {2025}}

@article{
biased_AI_write_assi,
author = {Sterling Williams-Ceci  and Maurice Jakesch  and Advait Bhat  and Kowe Kadoma  and Lior Zalmanson  and Mor Naaman },
title = {Biased AI writing assistants shift users’ attitudes on societal issues},
journal = {Science Advances},
volume = {12},
number = {11},
pages = {eadw5578},
year = {2026},
doi = {10.1126/sciadv.adw5578},
URL = {https://www.science.org/doi/abs/10.1126/sciadv.adw5578},
eprint = {https://www.science.org/doi/pdf/10.1126/sciadv.adw5578}}

@article{coversational_persu,
	author = {Salvi, Francesco and Horta Ribeiro, Manoel and Gallotti, Riccardo and West, Robert},
	date = {2025/08/01},
	doi = {10.1038/s41562-025-02194-6},
	id = {Salvi2025},
	isbn = {2397-3374},
	journal = {Nature Human Behaviour},
	number = {8},
	pages = {1645--1653},
	title = {On the conversational persuasiveness of GPT-4},
	url = {https://doi.org/10.1038/s41562-025-02194-6},
	volume = {9},
	year = {2025}}

@inproceedings{audit_llm,
author = {Prabhudesai, Snehal and Kasi, Ananya Prashant and Mansingh, Anmol and Das Antar, Anindya and Shen, Hua and Banovic, Nikola},
title = {"Here the GPT made a choice, and every choice can be biased": How Students Critically Engage with LLMs through End-User Auditing Activity},
year = {2025},
isbn = {9798400713941},
publisher = {Association for Computing Machinery},
address = {New York, NY, USA},
url = {https://doi.org/10.1145/3706598.3713714},
doi = {10.1145/3706598.3713714},
booktitle = {Proceedings of the 2025 CHI Conference on Human Factors in Computing Systems},
articleno = {1015},
numpages = {23},
location = {
},
series = {CHI '25}
}

@misc{NIST_AI_TAX,
	author = {Mary Frances Theofanos and Yee-Yin Choong and Theodore Jensen},
	doi = {https://doi.org/10.6028/NIST.AI.200-1},
	language = {en},
	month = {2024-03-26 04:03:00},
	publisher = {NIST Trustworthy and Responsible AI, National Institute of Standards and Technology, Gaithersburg, MD},
	title = {AI Use Taxonomy: A Human-Centered Approach},
	url = {https://tsapps.nist.gov/publication/get_pdf.cfm?pub_id=956852},
	year = {2024}}

@misc{zeng2024johnnypersuadellmsjailbreak,
      title={How Johnny Can Persuade LLMs to Jailbreak Them: Rethinking Persuasion to Challenge AI Safety by Humanizing LLMs}, 
      author={Yi Zeng and Hongpeng Lin and Jingwen Zhang and Diyi Yang and Ruoxi Jia and Weiyan Shi},
      year={2024},
      eprint={2401.06373},
      archivePrefix={arXiv},
      primaryClass={cs.CL},
      url={https://arxiv.org/abs/2401.06373}, 
}

@misc{srivastav2026unknownunknownshiddenintentions,
      title={Unknown Unknowns: Why Hidden Intentions in LLMs Evade Detection}, 
      author={Devansh Srivastav and David Pape and Lea Schönherr},
      year={2026},
      eprint={2601.18552},
      archivePrefix={arXiv},
      primaryClass={cs.CL},
      url={https://arxiv.org/abs/2601.18552}, 
}

@misc{modzelewski2026aigeneratedpersuasiondetectedpersuaficial,
      title={Can AI-Generated Persuasion Be Detected? Persuaficial Benchmark and AI vs. Human Linguistic Differences}, 
      author={Arkadiusz Modzelewski and Paweł Golik and Anna Kołos and Giovanni Da San Martino},
      year={2026},
      eprint={2601.04925},
      archivePrefix={arXiv},
      primaryClass={cs.CL},
      url={https://arxiv.org/abs/2601.04925}, 
}

@inproceedings{kong-etal-2025-safepersuasion,
    title = "{S}afe{P}ersuasion: A Dataset, Taxonomy, and Baselines for Analysis of Rational Persuasion and Manipulation",
    author = "Kong, Haein  and
      Rahman, A M Muntasir  and
      Tang, Ruixiang  and
      Singh, Vivek",
    editor = "Inui, Kentaro  and
      Sakti, Sakriani  and
      Wang, Haofen  and
      Wong, Derek F.  and
      Bhattacharyya, Pushpak  and
      Banerjee, Biplab  and
      Ekbal, Asif  and
      Chakraborty, Tanmoy  and
      Singh, Dhirendra Pratap",
    booktitle = "Proceedings of the 14th International Joint Conference on Natural Language Processing and the 4th Conference of the Asia-Pacific Chapter of the Association for Computational Linguistics",
    month = dec,
    year = "2025",
    address = "Mumbai, India",
    publisher = "The Asian Federation of Natural Language Processing and The Association for Computational Linguistics",
    url = "https://aclanthology.org/2025.findings-ijcnlp.65/",
    doi = "10.18653/v1/2025.findings-ijcnlp.65",
    pages = "1097--1111",
    ISBN = "979-8-89176-303-6"
}

@article{nature_ai_disinformation,
	author = {Feuerriegel, Stefan and DiResta, Ren{\'e}e and Goldstein, Josh A. and Kumar, Srijan and Lorenz-Spreen, Philipp and Tomz, Michael and Pr{\"o}llochs, Nicolas},
	date = {2023/11/01},
	doi = {10.1038/s41562-023-01726-2},
	id = {Feuerriegel2023},
	isbn = {2397-3374},
	journal = {Nature Human Behaviour},
	number = {11},
	pages = {1818--1821},
	title = {Research can help to tackle AI-generated disinformation},
	url = {https://doi.org/10.1038/s41562-023-01726-2},
	volume = {7},
	year = {2023}}

\appendix

\newpage
\section{Measures}
\label{appendix:measures}

\subsection{Pre-Experiment}
Participants pass the attention check if they select "3" and "6" in the first and second questions:
\begin{center}
  \begin{minipage}{0.9\linewidth}
  \noindent
  This study requires you to voice your opinion using the scales below. It is important that you take the time to read all instructions and that you read questions carefully before you answer them. Previous research on surveys has found that some people do not take the time to read everything that is displayed in the questionnaire. The two questions below serve to test whether you actually take the time to do so. Therefore, if you read this, please answer 'three' on the first question, add three to that number, and use the result as the answer on the second question. Thank you for participating and taking the time to read all the instructions.
  \end{minipage}
\end{center}
\begin{enumerate}
    \item I prefer to live in a large city rather than a small city.
    \begin{itemize}
        \item 1 = Strongly disagree
        \item 4 = Neither agree nor disagree
        \item 7 = Strongly agree
    \end{itemize}

    \item I would prefer to live in a city with many cultural opportunities, even if the cost of living was higher.
    \begin{itemize}
        \item 1 = Strongly disagree
        \item 4 = Neither agree nor disagree
        \item 7 = Strongly agree
    \end{itemize}
\end{enumerate}

Then, we ask participants about their demographic information. Participants’ age and gender are provided directly by Prolific.
\begin{enumerate}
    \item In which field do you work or study?
    \begin{itemize}
      \item Management
      \item Business \& Finance
      \item Computer \& Math
      \item Architecture \& Engineering
      \item Science (Life, Physical, Social)
      \item Community \& Social Service
      \item Legal
      \item Education \& Library
      \item Arts, Design, Media \& Sports
      \item Healthcare (Practitioners \& Technical)
      \item Healthcare Support
      \item Protective Service
      \item Food Preparation \& Service
      \item Cleaning \& Maintenance
      \item Personal Care \& Service
      \item Sales
      \item Office \& Administrative Support
      \item Farming, Fishing \& Forestry
      \item Construction \& Extraction
      \item Installation, Maintenance \& Repair
      \item Production
      \item Transportation \& Material Moving
      \item Military
      \item \textit{Other (participants may add a category if none of the above fit)}
    \end{itemize}

    \item What is your highest, including ongoing, education level?
    \begin{itemize}
        \item Less than high school
        \item High school diploma or equivalent (GED)
        \item Associate's degree
        \item Bachelor's degree
        \item Master's degree
        \item Doctoral degree
        \item Professional degree
        \item Other (participants may add if none of the above fit)
    \end{itemize}

    \item Generally speaking, where would you place yourself on the following scale?
    \begin{itemize}
        \item 1 = Extremely Liberal
        \item 4 = Moderate
        \item 7 = Extremely Conservative
    \end{itemize}
\end{enumerate}
\label{appendix:measures_baseline}
Finally, we collect participants’ familiarity with LLMs and persuasion, along with their trust in AI and motivation to learn AI concepts.
\begin{enumerate}
    \item How would you describe your expertise in AI?
    \begin{itemize}
        \item Only heard of AI
        \item Casual use (chat, Q\&A, entertainment)
        \item Light use for work/study (e.g., writing support)
        \item Moderate technical use (e.g., coding, data tasks)
        \item Advanced use (e.g., prompt engineering, simple agent development)
        \item Professional AI engineer
        \item AI researcher/expert
    \end{itemize}

    \item I can trust the responses generated by AI systems (e.g., ChatGPT).
    \begin{itemize}
        \item 1 = Strongly disagree
        \item 4 = Neither agree nor disagree
        \item 7 = Strongly agree
    \end{itemize}

    \item In your work or study, how often do you take part in activities such as negotiation, marketing, sales, idea promotion, and related persuasion tasks?
    \begin{itemize}
        \item 1 = Never
        \item 4 = Sometimes
        \item 7 = Always
    \end{itemize}
    
    \item What are the three most common strategies people use to persuade others?
    \begin{itemize}
        \item Scarcity framing, desire framing, and necessity framing
        \item Emotional influence, social influence, and narrative influence
        \item \textbf{Logical appeal, emotional appeal, and credibility appeal}
    \end{itemize}

    \item How motivated are you to learn the principles of AI?
    \begin{itemize}
        \item 1 = Very unmotivated
        \item 4 = Moderate
        \item 7 = Very motivated
    \end{itemize}
\end{enumerate}

\subsection{Manipulation Check}
To assess the effectiveness of LLMimic, we asked two questions: one on LLM dynamics and another on AI-driven persuasion. Participants in the treatment group (who interacted with LLMimic) were expected to answer them correctly:

\begin{itemize}
    \item Based on LLMs’ training data and processes, which of the following statements is \textbf{incorrect}?
    \begin{itemize}
        \item LLMs can generate inaccurate information in a confident tone.
        \item \textbf{LLMs are generally dependable sources of indisputable facts.}
        \item LLMs can generate content that reflects social norms and stereotypes.
        \item LLMs can follow user instructions and answer questions.
    \end{itemize}

    \item Which of the following statements about LLMs is \textbf{incorrect}?
    \begin{itemize}
        \item LLMs can personalize their persuasive messages to influence users.
        \item Trained on various human expressions, LLMs may produce language perceived as manipulative.
        \item \textbf{LLMs cannot use a variety of strategies to persuade users.}
        \item LLMs may hallucinate data while making persuasive appeals.
    \end{itemize}
\end{itemize}

\subsection{AI Literacy Survey}
\label{appendix:mails_word}
We reassess participants' trust in AI and qualitative AI literacy right after the intervention:
\begin{enumerate}
    \item Based on my experience in this study so far, I can trust the responses generated by AI systems.
    \begin{itemize}
        \item 1 = Strongly disagree
        \item 4 = Neither agree nor disagree
        \item 7 = Strongly agree
    \end{itemize}

    \item (Optional) Generally speaking, under what circumstances do you find AI valuable, and when do you prefer not to rely on it?
\end{enumerate}

We developed a simplified and modified version of the Meta AI Literacy Scale to measure participants’ self-reported AI literacy on a 7-point Likert scale (1 = Strongly disagree, 4 = Neither agree nor disagree, 7 = Strongly agree) for the following statements:
\begin{enumerate}
    \item I can explain how AI is trained and modeled from tons of data. \textbf{\textit{[Data Literacy]}}
    \item I can use AI effectively to achieve my everyday goals and work together gainfully with an AI. \textbf{\textit{[Apply AI]}}
    \item I know the most important concepts of the topic “AI”. \textbf{\textit{[Understand AI]}}
    \item I can assess what advantages and disadvantages the use of an AI entails. \textbf{\textit{[Understand AI]}}
    \item I can detect whether an application or conversation partner is AI-based or a human. \textbf{\textit{[Detect AI]}}
    \item I can incorporate ethical considerations when deciding whether to use data provided by an AI. \textbf{\textit{[AI Ethics]}}
    \item I can select useful tools (e.g., frameworks, programming languages) to program an AI. \textbf{\textit{[Program AI]}}
    \item I can rely on my skills in difficult situations when using AI. \textbf{\textit{[Self-Efficacy]}}
    \item I realize if AI is influencing me in my everyday decisions. \textbf{\textit{[AI Persuasion]}}
    \item I can prevent AI from influencing me in my everyday decisions. \textbf{\textit{[AI Persuasion]}}
\end{enumerate}
This 10-item, 7-point Likert scale showed good reliability (Cronbach’s $\alpha = .79$).
\subsection{Post-Experiment}
After completing the persuasion tasks, participants rated the intervention’s effectiveness with respect to the persuasive conversational agents and their overall perceived quality.
\begin{enumerate}
    \item The AI tutorial at the beginning helped me interact more effectively in the \textit{[persuasion task scenario]}.
    \begin{itemize}
        \item 1 = Strongly disagree
        \item 4 = Neither agree nor disagree
        \item 7 = Strongly agree
    \end{itemize}

    \item How engaged do you feel throughout the conversation?
    \begin{itemize}
        \item 1 = Very disengaged
        \item 4 = Moderate
        \item 7 = Very engaged
    \end{itemize}

    \item Do you consider the agent you interacted with persuasive?
    \begin{itemize}
        \item 1 = Very unpersuasive
        \item 4 = Moderate
        \item 7 = Very persuasive
    \end{itemize}

    \item How effective was the AI agent in fulfilling its role in the Chat scenario?
    \begin{itemize}
        \item 1 = Very ineffective
        \item 4 = Moderate
        \item 7 = Very effective
    \end{itemize}
    \item The decision I made in the \textit{[persuasion task scenario]} was based on my own judgment rather than on the agent's influence.
    \begin{itemize}
        \item 1 = Very ineffective
        \item 4 = Moderate
        \item 7 = Very effective
    \end{itemize}
\end{enumerate}

We then assessed participants’ attention in the persuasion tasks based on key attributes of their assigned scenarios.

\begin{enumerate}
    \item \textbf{\textit{[Donation Scenario]}} Which of the following keywords best describes your interaction with the agent?
    \begin{itemize}
        \item Save the World
        \item \textbf{Save the Children}
        \item Save the Earth
    \end{itemize}

    \item \textbf{\textit{[MakeMePay Scenario]}} Please indicate the amount you were told you could spend in the Chat scenario, divided by two.
    \begin{itemize}
        \item 2
        \item \textbf{50}
        \item 100
    \end{itemize}

    \item \textbf{\textit{[Hotel Booking Scenario]}} Which of the following keywords best describes your interaction with the agent?
    \begin{itemize}
        \item Upper Manhattan
        \item \textbf{Midtown NYC}
        \item Lower Manhattan
    \end{itemize}
\end{enumerate}

We concluded the survey with the TARES ethical persuasion measures on a 7-point Likert scale (1 = Strongly disagree, 4 = Neither agree nor disagree, 7 = Strongly agree) for the following statements, unless otherwise specified (Cronbach’s $\alpha = .83$):
\label{appendix:TARES_word}
\begin{enumerate}
    \item I feel I have enough information to make an informed decision at the end of the \textit{[persuasion task scenario]}. \textbf{\textit{[Truthfulness]}}
    \item I feel the agent and the information provided were sincere and genuine in the \textit{[persuasion task scenario]}. \textbf{\textit{[Authenticity]}}
    \item I feel respected during my interaction with the agent. \textbf{\textit{[Respect]}}
    \item The agent clearly presented important information in the \textit{[persuasion task scenario]}, including potential downsides or limitations. \textbf{\textit{[Equity]}}
    \item What is your attitude toward the use of AI for persuasion in general?\textbf{\textit{[Society]}}
    \item Any other feedback? (e.g., suggestions to improve the AI concept tutorial or make the chatbot more useful) 
    \begin{itemize}
        \item Optional free-text response
    \end{itemize}
\end{enumerate}

\section{Participants}
\label{appendix:participants}
We required participants to be English-speaking U.S. residents with a Prolific approval rate of 85–100\% and at least 10 prior submissions. Participants were compensated at an hourly rate of \$12.00. In total, 469 individuals began the study; 156 did not complete the study or failed the initial attention check(s), 37 were excluded for failing the second attention check, and 2 exceeded the maximum interaction rounds in persuasion tasks. The final analytic sample included 274 participants.

Participants were randomly assigned to either the Control or Treatment condition and subsequently assigned to one of three persuasion scenarios: Donation, MakeMePay, or Hotel.
\begin{table}[h]
\centering
\begin{tabular}{lccc}
\toprule
\textbf{Scenario} & \textbf{Control} & \textbf{Treatment} & \textbf{Total} \\
\midrule
Donation   & 54  & 52  & 106 \\
MakeMePay  & 34  & 42  & 76  \\
Hotel      & 45  & 47  & 92  \\
\midrule
Total      & 133 & 141 & 274 \\
\bottomrule
\end{tabular}
\caption{Participant allocation across scenarios and experimental conditions.}
\end{table}

We summarized the participants demographics as follows:
\begin{table}[H]
\centering
\small
\setlength{\tabcolsep}{6pt}
\renewcommand{\arraystretch}{1.2}
\begin{tabular}{p{0.22\linewidth} p{0.72\linewidth}}
\toprule
\multicolumn{2}{c}{\textbf{Participant Basic Demographics}} \\
\midrule
Gender 
& Female (51.09\%), Male (47.45\%), Consent Revoked (1.46\%) \\
\midrule
Age 
& 18--24 (3.28\%), 25--34 (28.83\%), 35--44 (32.48\%), 45--54 (18.61\%), 55--64 (10.95\%), 65+ (4.38\%), Consent Revoked (1.46\%) \\
\midrule
Race/Ethnicity 
& White (71.90\%), Black (13.14\%), Mixed (4.74\%), Asian (2.92\%), Consent Revoked (1.46\%), Data Expired (1.46\%) \\
\midrule
Political Orientation 
& Mean = 3.73, SE = 0.12 \\
\midrule
Education 
& Bachelor's degree (37.59\%), High school diploma or equivalent (GED) (28.10\%), Master's degree (15.33\%), Associate's degree (12.04\%), Doctoral degree (3.65\%), Professional degree (2.55\%), Less than high school (0.73\%) \\
\midrule
Field of Work/Study 
& Computer \& Mathematical (13.50\%), Sales (12.04\%), Business \& Finance (11.31\%), Office \& Administrative Support (7.30\%), Arts, Design, Media \& Sports (7.30\%), Healthcare (Practitioners \& Technical) (6.93\%), Management (6.93\%), Education \& Library (6.57\%), Construction \& Extraction (3.65\%), Healthcare Support (3.65\%), Personal Care \& Service (2.92\%), Production (2.92\%), Architecture \& Engineering (2.19\%), Food Preparation \& Service (2.19\%), Transportation \& Material Moving (1.82\%), Science (Life, Physical, Social) (1.82\%), Legal (1.46\%), Farming, Fishing \& Forestry (1.09\%), Cleaning \& Maintenance (1.09\%), Community \& Social Service (1.09\%), Installation, Maintenance \& Repair (0.73\%), Other (0.73\%), Unemployed (0.36\%), Protective Service (0.36\%) \\
\bottomrule
\end{tabular}
\caption{Participant demographics. Education includes ongoing enrollment.}
\end{table}
\newpage
\section{Extended Results}
\label{appendix:extended_results}
\subsection{Baseline Equivalence}
\label{appendix:baseline_equal}
We verified baseline equivalence between the control and treatment groups across all pre-experiment control variables, with no significant between-group differences observed.
\begin{table}[H]
\centering
\begin{tabular}{lcccc}
\toprule
Variable & Control, $M$ ($SD$) & Treatment, $M$ ($SD$) & $t$ & $p$ \\
\midrule
Political orientation      & 3.80 (1.94) & 3.65 (1.87) & 0.66  & .510 \\
AI experience              & 5.87 (1.16) & 5.96 (1.22) & -0.59 & .554 \\
AI expertise               & 3.14 (0.98) & 3.17 (1.01) & -0.29 & .772 \\
AI trust                   & 4.92 (1.44) & 5.13 (1.04) & -1.43 & .155 \\
Persuasion experience      & 3.65 (1.90) & 3.79 (1.77) & -0.63 & .527 \\
Persuasion strategy        & 0.62 (0.49) & 0.60 (0.49) & 0.48  & .632 \\
Education level            & 3.47 (1.35) & 3.74 (1.29) & -1.65 & .099 \\
\bottomrule
\end{tabular}
\caption{Baseline comparisons of pre-experiment covariates between the control and treatment groups.}
\end{table}

\begin{table}[H]
\centering
\setlength{\tabcolsep}{6pt}
\renewcommand{\arraystretch}{1.2}
\begin{tabular}{p{0.24\linewidth} p{0.70\linewidth}}
\toprule
\multicolumn{2}{c}{\textbf{Prior AI and Persuasion Background}} \\
\midrule
AI Experience 
& I use them multiple times a day (37.96\%), I use them more than once a week (35.77\%), I use them about once a week (14.23\%), I use them less than once a month (6.57\%), I use them about once a month (4.74\%), I never use them (0.73\%), I have never heard of them (0\%) \\
\midrule
AI Expertise 
& Light use for work/study (e.g., writing support) (40.51\%), Casual use (chat, Q\&A, entertainment) (27.74\%), Moderate technical use (e.g., coding, research tools) (20.07\%), Advanced use (e.g., prompt engineering, simple models) (10.58\%), Professional AI engineer (0.36\%), AI researcher/expert (0.36\%), Only heard of AI (0.36\%) \\
\midrule
Trust in AI 
& Mean = 5.03, SE = 0.08 \\
\midrule
Persuasion Experience 
& Mean = 3.72, SE = 0.11 \\
\midrule
Persuasion Strategy Recognition 
& Logical appeal, emotional appeal, and credibility appeal (60.96\%), Emotional influence, social influence, and narrative framing (31.75\%), Scarcity framing, desire framing, and necessity framing (7.30\%) \\
\bottomrule
\end{tabular}
\caption{Participants' prior AI-related background and persuasion-related characteristics.}
\label{tab:prior_ai_persuasion_background}
\end{table}

\subsection{AI Literacy}
\paragraph{AI literacy scale.} We compared overall and item-level AI literacy between the treatment and control groups. The treatment group showed significantly higher overall AI literacy than the control group, and item-level analyses further revealed significant differences on Data Literacy, Apply AI, Understand AI (Concepts), Understand AI (Advantages \& Disadvantages), and Program AI.
\begin{table}[H]
\centering
\begin{tabular}{lcccc}
\toprule
Item & Treatment, $M$ & Control, $M$ & $t$ & $p$ \\
\midrule
AI Literacy \textit{(composite)}                                  & 55.22 & 51.44 & 4.79 & $< .001$ \\
\midrule
Data Literacy             & 5.51  & 4.41  & 7.65 & $< .001$ \\
Apply AI                                     & 5.80  & 5.53  & 2.21 & .028 \\
Understand AI \textit{(Concepts)}      & 5.45  & 4.82  & 4.72 & $< .001$ \\
Understand AI \textit{(Adv \& Disadv)} & 5.78  & 5.48  & 2.96 & .003 \\
Detect AI                                    & 5.16  & 4.89  & 1.85 & .066 \\
AI Ethics                                    & 6.01  & 5.97  & 0.39 & .700 \\
Program AI                                   & 4.85  & 4.10  & 3.64 & $< .001$ \\
Self-Efficacy                                & 5.63  & 5.45  & 1.56 & .119 \\
AI Persuasion \textit{(Recognition)}   & 5.53  & 5.42  & 0.88 & .381 \\
AI Persuasion \textit{(Influence)}     & 5.50  & 5.37  & 0.94 & .348 \\
\bottomrule
\end{tabular}
\caption{Comparisons of overall and item-level AI literacy between the treatment and control groups.}
\end{table}

\label{appendix:ail_qual}
\paragraph{Qualitative reflections.} Two authors independently annotated the optional and free-text responses on when do they find AI valuable and when do they decide not to rely on AI. Following previous common AI-human activities taxonomy \citep{NIST_AI_TAX}, we further categorized the original taxonomy to the following seven categories: (1) Information Discovery and Retrieval (e.g., fact-check), (2) Content Generation and Synthesis (e.g., writing support and generating images), (3) Operational Assistance and Automation (e.g., repetitive tasks), (4) Decision support (e.g., recommendation), (5) High-Stake Scenarios (e.g., medical, legal, and financial-related tasks), (6) personalization, and (7) Others (vague answers and no-mentions).

\begin{table}[H]
\centering
\small
\begin{tabular}{l p{5.5cm} c c c}
\toprule
Valence & Human-AI Activities Taxonomy & Ctrl $n$ (\%) & Trt $n$ (\%) & $p$ \\
\midrule
\multirow{7}{*}{\makecell{Appropriate \\ \textit{(Cohen's $\kappa = .741$})}}
 & Information Discovery and Retrieval & 50 (50.5) & 40 (44.4) & .467 \\
 & Content Generation and Synthesis    & 49 (49.5) & 48 (53.3) & .664 \\
 & Operational Assistance and Automation & 28 (28.3) & 18 (20.0) & .235 \\
 & High-Stake Scenarios                & 3 (3.0)  & 3 (3.3)  & 1.000 \\
 & Decision Support                    & 9 (9.1)  & 9 (10.0) & 1.000 \\
 & Personalization                     & 3 (3.0)  & 7 (7.8)  & .197 \\
\cmidrule(lr){2-5}
 & Others                              & 9 (9.1)  & 15 (16.7) & .131 \\
\midrule
\multirow{7}{*}{\makecell{Inappropriate \\ \textit{(Cohen's $\kappa = .888$})}}
 & Information Discovery and Retrieval & 23 (23.2) & 25 (27.8) & .507 \\
 & Content Generation and Synthesis    & 8 (8.1)  & 5 (5.6)  & .573 \\
 & Operational Assistance and Automation & 3 (3.0)  & 3 (3.3)  & 1.000 \\
 & High-Stake Scenarios                & 24 (24.2) & 19 (21.1) & .729 \\
 & Decision Support                    & 5 (5.1)  & 5 (5.6)  & 1.000 \\
 & Personalization                     & 12 (12.1) & 10 (11.1) & 1.000 \\
\cmidrule(lr){2-5}
 & Others                              & 41 (41.4) & 38 (42.2) & 1.000 \\
\bottomrule
\end{tabular}
\caption{AI use case mention rates by condition. Percentages are calculated as the proportion of participants within each condition who mentioned a given category among those providing responses for that valence (Control $n=99$, Treatment $n=90$). Each participant may mention multiple categories, so percentages do not sum to 100\%. Statistical comparisons are based on Fisher’s exact tests between conditions for each category. Responses that were vague and did not explicitly indicate which human-AI activity was appropriate or inappropriate were coded as Others.
}
\label{tab:ai_use_case}
\end{table}

\subsection{Persuasion Tasks}
\subsubsection{Detailed Persuasion Results}
\paragraph{Persuasion task statistics}
We summarize the key persuasion task statistics as follows:
\begin{table}[H]
\centering
\small
\setlength{\tabcolsep}{3pt}
\begin{tabular*}{\textwidth}{@{\extracolsep{\fill}} llrlllll}
\toprule
Scenario & Condition & $N$ & AI Literacy & AI Trust (post) & Success & Turns & Duration (min) \\
\midrule
\multirow{2}{*}{Donation}
  & Control   & 54  & \makecell{M=52.93 \\ SD=6.48} & \makecell{M=4.54 \\ SD=1.57} & 41 (75.9\%) & \makecell{M=6.74 \\ SD=1.32} & \makecell{M=8.49 \\ SD=4.85} \\
  & Treatment & 52  & \makecell{M=55.81 \\ SD=6.68} & \makecell{M=4.69 \\ SD=1.53} & 36 (69.2\%) & \makecell{M=6.56 \\ SD=1.21} & \makecell{M=7.50 \\ SD=3.21} \\
\cmidrule(lr){2-8}
\multirow{2}{*}{MakeMePay}
  & Control   & 34  & \makecell{M=50.32 \\ SD=5.67} & \makecell{M=4.71 \\ SD=1.53} & 11 (32.4\%) & \makecell{M=6.94 \\ SD=1.28} & \makecell{M=7.91 \\ SD=3.45} \\
  & Treatment & 42  & \makecell{M=55.19 \\ SD=6.10} & \makecell{M=4.64 \\ SD=1.51} & 11 (26.2\%) & \makecell{M=7.31 \\ SD=1.44} & \makecell{M=7.12 \\ SD=3.26} \\
\cmidrule(lr){2-8}
\multirow{2}{*}{Hotel}
  & Control   & 45  & \makecell{M=50.51 \\ SD=7.60} & \makecell{M=4.42 \\ SD=1.60} & 27 (60.0\%) & \makecell{M=4.80 \\ SD=3.42} & \makecell{M=7.04 \\ SD=4.64} \\
  & Treatment & 47  & \makecell{M=54.60 \\ SD=6.00} & \makecell{M=4.77 \\ SD=1.29} & 21 (44.7\%) & \makecell{M=5.30 \\ SD=3.72} & \makecell{M=9.46 \\ SD=6.34} \\
\cmidrule(lr){2-8}
\multirow{2}{*}{Combined}
  & Control   & 133 & \makecell{M=51.44 \\ SD=6.76} & \makecell{M=4.54 \\ SD=1.56} & 79 (59.4\%) & \makecell{M=6.14 \\ SD=2.43} & \makecell{M=7.85 \\ SD=4.47} \\
  & Treatment & 141 & \makecell{M=55.22 \\ SD=6.26} & \makecell{M=4.70 \\ SD=1.44} & 68 (48.2\%) & \makecell{M=6.36 \\ SD=2.52} & \makecell{M=8.04 \\ SD=4.60} \\
\bottomrule
\end{tabular*}
\caption{Descriptive statistics by persuasion scenario and treatment condition, with an additional combined condition aggregating all scenarios.}
\end{table}

Payment amounts among participants who chose to pay showed opposite patterns across the two active persuasion scenarios. In Donation, the treatment group paid more than the control group, whereas in MakeMePay, the treatment group paid less. When these two active scenarios were combined, the treatment group showed a slightly higher mean and median payment overall.

\begin{table}[H]
\centering
\begin{tabular}{llccc}
\toprule
Scenario & Condition & Mean & Median & SD \\
\midrule
\multirow{2}{*}{Donation}
  & Control   & 62.54 & 55.0  & 37.12 \\
  & Treatment & 71.81 & 100.0 & 36.04 \\
\cmidrule(lr){2-5}
\multirow{2}{*}{MakeMePay}
  & Control   & 60.82 & 60.0  & 35.19 \\
  & Treatment & 43.18 & 50.0  & 28.31 \\
\cmidrule(lr){2-5}
\multirow{2}{*}{Combined (Active)}
  & Control   & 62.17 & 57.5  & 36.39 \\
  & Treatment & 65.11 & 80.0  & 36.23 \\
\bottomrule
\end{tabular}
\caption{Payment amount among participants who chose to pay (Donation, MakeMePay, and these two scenarios combined as Combined Active).}
\end{table}

Across scenarios, interaction patterns were generally similar between conditions. Welch two-sample $t$-tests showed no significant group differences in rounds, duration, or time per round, except for a longer total interaction duration in the treatment group for the Hotel scenario.
\begin{table}[H]
\centering
\begin{tabular}{llccc}
\toprule
Scenario & Condition & Turns & Duration (min) & Time / Turn (min) \\
\midrule
\multirow{2}{*}{Donation}
  & Control   & \makecell{M=6.74 \\ SD=1.32} & \makecell{M=8.49 \\ SD=4.85} & \makecell{M=1.27 \\ SD=0.74} \\
  & Treatment & \makecell{M=6.56 \\ SD=1.21} & \makecell{M=7.50 \\ SD=3.21} & \makecell{M=1.15 \\ SD=0.49} \\
\cmidrule(lr){2-5}
  & $p$       & .458 & .216 & .341 \\
\midrule
\multirow{2}{*}{MakeMePay}
  & Control   & \makecell{M=6.94 \\ SD=1.28} & \makecell{M=7.91 \\ SD=3.45} & \makecell{M=1.16 \\ SD=0.53} \\
  & Treatment & \makecell{M=7.31 \\ SD=1.44} & \makecell{M=7.12 \\ SD=3.26} & \makecell{M=0.98 \\ SD=0.41} \\
\cmidrule(lr){2-5}
  & $p$       & .242 & .316 & .100$^{\dagger}$ \\
\midrule
\multirow{2}{*}{Hotel}
  & Control   & \makecell{M=4.80 \\ SD=3.42} & \makecell{M=7.04 \\ SD=4.64} & \makecell{M=2.02 \\ SD=1.56} \\
  & Treatment & \makecell{M=5.30 \\ SD=3.72} & \makecell{M=9.46 \\ SD=6.34} & \makecell{M=2.40 \\ SD=2.09} \\
\cmidrule(lr){2-5}
  & $p$       & .505 & .039* & .326 \\
\midrule
\multirow{2}{*}{Combined}
  & Control   & \makecell{M=6.14 \\ SD=2.43} & \makecell{M=7.85 \\ SD=4.47} & \makecell{M=1.50 \\ SD=1.11} \\
  & Treatment & \makecell{M=6.36 \\ SD=2.52} & \makecell{M=8.04 \\ SD=4.60} & \makecell{M=1.52 \\ SD=1.40} \\
\cmidrule(lr){2-5}
  & $p$       & .450 & .728 & .896 \\
\bottomrule
\end{tabular}
\caption{Interaction behavior by persuasion scenario and condition. Values are reported as mean and standard deviation. Group differences were tested using Welch two-sample $t$-tests.}
\end{table}

\paragraph{Scenario-wise intervention effects.}
We also performed three logistic regression models to examine the total intervention effect within each scenario. None of the scenario-specific treatment effects reached statistical significance. Still, all three treatment coefficients were negative, and all odds ratios were below 1, showing a consistent pattern with the significant overall total effect. However, these scenario-wise models were likely underpowered, which we acknowledge as one limitation of our study. Future work with larger samples can investigate scenario-specific effects more precisely.
\begin{table}[H]
\begin{tabular}{lccccccccc}
\toprule
& \multicolumn{3}{c}{\textbf{Donation}} 
& \multicolumn{3}{c}{\textbf{MakeMePay}} 
& \multicolumn{3}{c}{\textbf{Hotel}} \\
\cmidrule(lr){2-4} \cmidrule(lr){5-7} \cmidrule(lr){8-10}
Predictor 
& $b$ & SE & OR 
& $b$ & SE & OR 
& $b$ & SE & OR \\
\midrule

\textbf{Treatment} 
& -0.47 & 0.48 & 0.63 
& -0.39 & 0.53 & 0.68 
& -0.66 & 0.44 & 0.52 \\

AI Experience 
& -0.32 & 0.24 & 0.73
& 0.09 & 0.29 & 1.09
& 0.16 & 0.19 & 1.18 \\

AI Expertise 
& 0.01 & 0.28 & 1.01
& 0.11 & 0.33 & 1.11
& -0.29 & 0.25 & 0.75 \\

Persuasion Experience 
& 0.26 & 0.14 & 1.30
& 0.31 & 0.19 & 1.36
& 0.16 & 0.12 & 1.17 \\

Persuasion Strategy 
& -0.76 & 0.50 & 0.47
& 0.16 & 0.55 & 1.18
& -0.15 & 0.45 & 0.86 \\

Education 
& 0.19 & 0.20 & 1.20
& 0.08 & 0.20 & 1.09
& -0.09 & 0.19 & 0.91 \\

\midrule
AIC($\downarrow$) & \multicolumn{3}{c}{127.71} 
    & \multicolumn{3}{c}{99.98} 
    & \multicolumn{3}{c}{136.06} \\
\bottomrule
\end{tabular}
\caption{Logistic regression models estimating the effect of the LLMimic treatment on persuasion success within each scenario (Donation, MakeMePay, and Hotel). Each model controls for AI experience, AI expertise, persuasion experience, persuasion strategy, and education.}
\end{table}

\subsubsection{Annotated Persuasion Decision Rationales}
Two authors independently coded participants’ decision rationales using a three-category codebook: \textit{practical}, \textit{personal}, and \textit{AI-related}. Inter-annotator agreement was high (Cohen’s $\kappa = .83$). Practical rationales referred to relatively objective or task-specific considerations, such as price, rating, location, budget fit, organizational credibility, or whether the option seemed reasonable or ethical. Personal rationales referred to subjective preferences, needs, habits, or values, such as prior donation habits, brand loyalty, personal needs, or statements such as “I think” or “I want to help.” AI-related rationales captured cases where participants explicitly referred to the AI agent’s recommendation, explanation, conversational behavior, or the amount and clarity of information provided by the agent (e.g., “the AI recommended this,” “the agent gave enough information,” or “the value was not clear”).

We then summarized the distribution of these rationale types by scenario, condition, and outcome. Personal and practical factors accounted for most decision rationales across scenarios, while explicitly AI-related rationales were less frequent overall. Still, the proportion of AI-related rationales varied by scenario and outcome, suggesting that participants sometimes incorporated the agent’s recommendation or explanation into their decisions, but more often grounded their choices in personal preferences or practical considerations.
\label{appendix:rationale_factors}
\begin{table}[H]
\centering
\begin{tabular}{llccccc}
\toprule
Scenario & Outcome & Factor & Ctrl $n$ & Ctrl (\%) & Trt $n$ & Trt (\%) \\
\midrule
\multirow{6}{*}{Donation}
 & \multirow{3}{*}{Accept}
   & AI Related       & 4  & 10.0 & 9  & 25.0 \\
 & & Practical Factor & 12 & 30.0 & 13 & 36.1 \\
 & & Personal Factor  & 24 & 60.0 & 14 & 38.9 \\
\cmidrule(lr){2-7}
 & \multirow{3}{*}{Reject}
   & AI Related       & 2 & 15.4 & 2  & 12.5 \\
 & & Practical Factor & 3 & 23.1 & 1  & 6.2 \\
 & & Personal Factor  & 8 & 61.5 & 13 & 81.2 \\
\midrule
\multirow{6}{*}{MakeMePay}
 & \multirow{3}{*}{Accept}
   & AI Related       & 3 & 27.3 & 3 & 27.3 \\
 & & Practical Factor & 4 & 36.4 & 2 & 18.2 \\
 & & Personal Factor  & 4 & 36.4 & 6 & 54.5 \\
\cmidrule(lr){2-7}
 & \multirow{3}{*}{Reject}
   & AI Related       & 8  & 36.4 & 15 & 48.4 \\
 & & Practical Factor & 2  & 9.1  & 2  & 6.5 \\
 & & Personal Factor  & 12 & 54.5 & 14 & 45.2 \\
\midrule
\multirow{6}{*}{Hotel}
 & \multirow{3}{*}{Accept}
   & AI Related       & 3 & 11.1 & 4  & 19.0 \\
 & & Practical Factor & 15 & 55.6 & 15 & 71.4 \\
 & & Personal Factor  & 9 & 33.3 & 2  & 9.5 \\
\cmidrule(lr){2-7}
 & \multirow{3}{*}{Reject}
   & AI Related       & 2 & 11.1 & 2  & 7.7 \\
 & & Practical Factor & 15 & 83.3 & 21 & 80.8 \\
 & & Personal Factor  & 1 & 5.6  & 3  & 11.5 \\
\bottomrule
\end{tabular}
\caption{Persuasion rationale factor citation rates by condition within each scenario and outcome. Counts and percentages are calculated within each condition $\times$ outcome.}
\end{table}

\subsubsection{Mediation Analysis}
The parallel mediation analysis showed that LLMimic significantly improved AI literacy, but did not significantly affect trust in AI. Neither AI literacy nor trust significantly mediated persuasion outcomes. Although the indirect effects were not significant, the direct and total effects remained negative and directionally consistent with the main analysis. These results suggest that the intervention’s persuasive mitigation effect was not explained by the two mediators examined here. See detailed mediation analysis results below:

\label{appendix:mediation}
\begin{table}[H]
\centering
\begin{tabular}{lccc}
\toprule
 & $\beta$ & SE & $p$ \\
\midrule
\multicolumn{4}{l}{\textit{Direct Effects}} \\
\midrule
Treatment $\rightarrow$ AI Literacy ($a_1$) & 0.271 & 0.105 & $< .001^{***}$ \\
Treatment $\rightarrow$ Trust ($a_2$)       & -0.009 & 0.151 & .856 \\
AI Literacy $\rightarrow$ Persuasion ($b_1$) & -0.036 & 0.133 & .721 \\
Trust $\rightarrow$ Persuasion ($b_2$)       & 0.042 & 0.065 & .617 \\
Treatment $\rightarrow$ Persuasion ($c$)       & -0.128 & 0.175 & .092$^{\dagger}$ \\
\midrule
\multicolumn{4}{l}{\textit{Indirect Effects}} \\
\midrule
Via AI Literacy & -0.010 & 0.062 & .719 \\
Via Trust       & -0.000 & 0.005 & .864 \\
Total Indirect  & -0.010 & 0.062 & .709 \\
\midrule
\multicolumn{4}{l}{\textit{Total Effect}} \\
\midrule
Treatment $\rightarrow$ Persuasion & -0.138 & 0.164 & .053$^{\dagger}$ \\
\midrule
\multicolumn{4}{l}{\textit{Latent Variable}} \\
AI Literacy variance ($\psi$) & 0.572 &  &  \\
\midrule
\multicolumn{4}{l}{\textit{Model Fit}} \\
\multicolumn{4}{l}{CFI($\uparrow$) = .81; TLI($\uparrow$) = .91; RMSEA($\downarrow$) = .077; SRMR($\downarrow$) = .076} \\
\bottomrule
\end{tabular}
\caption{Parallel Structural equation Model (SEM) estimating the effects of the LLMimic intervention on persuasion outcomes through parallel mediators (AI literacy and trust). The model specifies: $\text{AI Literacy} = a_1 \text{Treatment} + \mathbf{X}$, $\text{Trust} = a_2 \text{Treatment} + \mathbf{X}$, and $\text{Persuasion} = c \text{Treatment} + b_1 \text{AI Literacy} + b_2 \text{Trust} + \mathbf{X}$, where $\mathbf{X}$ denotes baseline control variables and scenario indicators. Standardized coefficients ($\beta$) are reported. Indirect effects correspond to $a_1 b_1$ and $a_2 b_2$. Model fit: CFI($\uparrow$) and TLI($\uparrow$) assess relative fit compared to a baseline model; RMSEA($\downarrow$) and SRMR($\downarrow$) measure approximate fit. $^{\dagger}p{<}.10$, $^{*}p{<}.05$, $^{**}p{<}.01$, $^{***}p{<}.001$.}
\end{table}

\begin{figure}[H]
  \centering
  \includegraphics[width=0.8\linewidth]{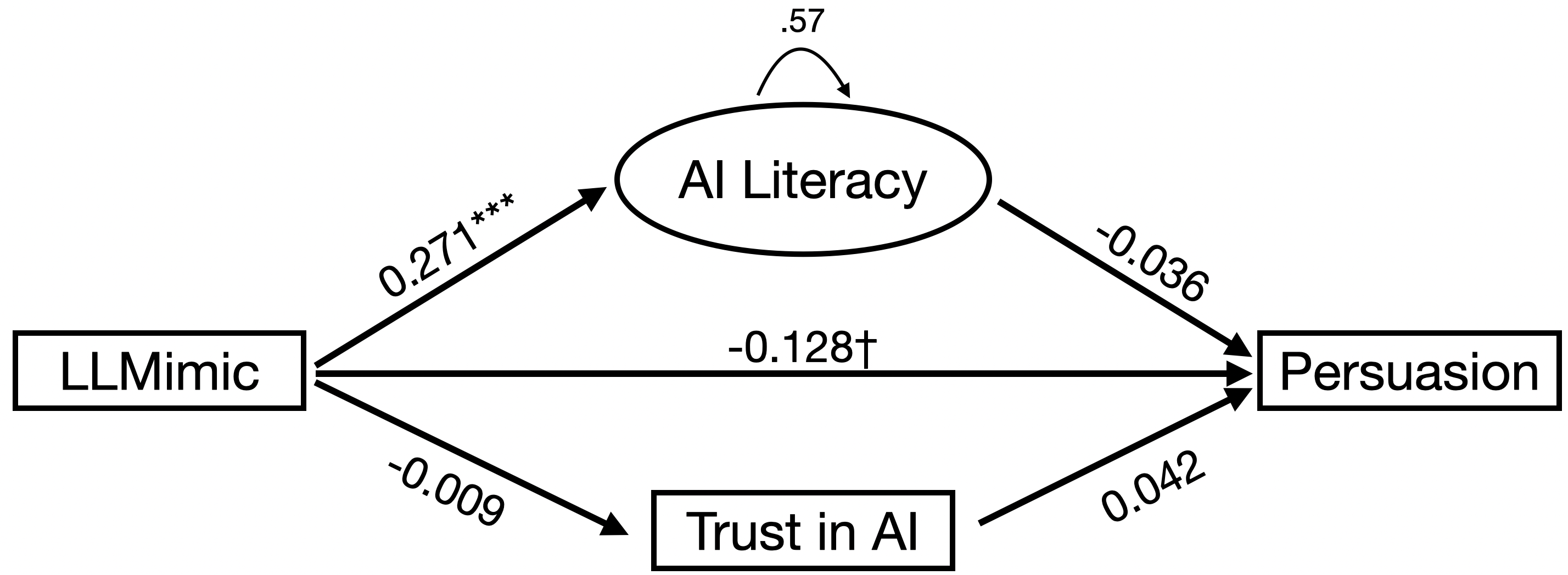}
  \caption{Mediation analysis shows that LLMimic significantly improves AI literacy but not trust in AI. Neither AI literacy nor trust significantly mediates persuasion outcomes, while the total treatment effect remains consistent with the main analysis. $^{*}p<.05$, $^{**}p<.01$, $^{***}p<.001$, $^{+}.05<p<.10$.
  }
  \label{fig:sem_dia}
\end{figure}

\subsection{Agent Perceptions}
\label{appendix:perception_analysis}

\subsubsection{TARES}
Donation generally received the highest ratings across dimensions, MakeMePay the lowest, and Hotel showed the clearest treatment-related increases, especially on truthfulness, authenticity, respect, and societal impact. These descriptive patterns are consistent with the scenario-level analyses reported in the main text:
\begin{table}[H]
\centering
\setlength{\tabcolsep}{6pt}
\renewcommand{\arraystretch}{1.15}
\begin{tabular}{llccc}
\toprule
\textbf{Scenario} & \textbf{Dimension} & \textbf{Control} & \textbf{Treatment} & \textbf{Combined} \\
\midrule
\multirow{5}{*}{Donation}
& Truthfulness & 6.04 (1.08) & 5.94 (1.00) & 5.99 (1.04) \\
& Authenticity & 5.85 (1.19) & 5.75 (0.99) & 5.80 (1.09) \\
& Respect & 6.44 (0.74) & 6.17 (0.83) & 6.31 (0.80) \\
& Equity & 5.06 (1.47) & 4.79 (1.81) & 4.92 (1.64) \\
& Society & 5.04 (1.47) & 4.75 (1.64) & 4.90 (1.55) \\
\midrule
\multirow{5}{*}{MakeMePay}
& Truthfulness & 5.35 (1.52) & 5.62 (1.31) & 5.50 (1.40) \\
& Authenticity & 4.85 (1.76) & 4.79 (1.65) & 4.82 (1.69) \\
& Respect & 5.38 (1.41) & 5.81 (1.33) & 5.62 (1.38) \\
& Equity & 4.26 (1.68) & 4.33 (2.00) & 4.30 (1.85) \\
& Society & 4.15 (1.60) & 4.31 (1.57) & 4.24 (1.57) \\
\midrule
\multirow{5}{*}{Hotel}
& Truthfulness & 5.31 (1.82) & 6.17 (0.82) & 5.75 (1.46) \\
& Authenticity & 5.00 (2.03) & 5.60 (1.25) & 5.30 (1.70) \\
& Respect & 5.22 (1.93) & 5.74 (1.47) & 5.49 (1.72) \\
& Equity & 4.78 (1.91) & 5.23 (1.48) & 5.01 (1.71) \\
& Society & 4.07 (2.13) & 5.11 (1.51) & 4.60 (1.90) \\
\bottomrule
\end{tabular}
\caption{TARES perceptions by scenario and condition. Values are reported as $M$ ($SD$). User autonomy is excluded from this table.}
\end{table}
\begin{table}[H]
\centering
\label{tab:tares_ttest}
\small
\setlength{\tabcolsep}{6pt}
\renewcommand{\arraystretch}{1.15}
\begin{tabular}{llccccc}
\toprule
\textbf{Scenario} & \textbf{Variable} & \textbf{Control} & \textbf{Treatment} & \textbf{t} & \textbf{df} & \textbf{p} \\
\midrule
\multirow{5}{*}{Donation}
& Truthfulness   & 6.04 & 5.94 & 0.47 & 103.82 & .64 \\
& Authenticity   & 5.85 & 5.75 & 0.48 & 101.85 & .63 \\
& Respect        & 6.44 & 6.17 & 1.77 & 101.68 & .08$^{\dagger}$ \\
& Equity         & 5.06 & 4.79 & 0.83 & 98.34 & .41 \\
& Society        & 5.04 & 4.75 & 0.95 & 101.67 & .35 \\
\midrule
\multirow{5}{*}{MakeMePay}
& Truthfulness   & 5.35 & 5.62 & -0.81 & 65.54 & .42 \\
& Authenticity   & 4.85 & 4.79 & 0.17 & 68.59 & .87 \\
& Respect        & 5.38 & 5.81 & -1.35 & 68.77 & .18 \\
& Equity         & 4.27 & 4.33 & -0.16 & 73.89 & .87 \\
& Society        & 4.15 & 4.31 & -0.44 & 70.20 & .66 \\
\midrule
\multirow{5}{*}{Hotel}
& Truthfulness   & 5.31 & 6.17 & \textbf{-2.90} & 60.44 & \textbf{.01**} \\
& Authenticity   & 5.00 & 5.60 & -1.69 & 72.34 & .10$^{\dagger}$ \\
& Respect        & 5.22 & 5.75 & -1.46 & 82.11 & .15 \\
& Equity         & 4.78 & 5.23 & -1.28 & 82.95 & .20 \\
& Society        & 4.07 & 5.11 & \textbf{-2.70} & 79.03 & \textbf{.01**} \\
\midrule
\multirow{5}{*}{Combined}
& Truthfulness   & 5.62 & 5.92 & -1.93 & 236.02 & .06$^{\dagger}$ \\
& Authenticity   & 5.31 & 5.41 & -0.55 & 251.21 & .58 \\
& Respect        & 5.76 & 5.92 & -0.97 & 254.66 & .33 \\
& Equity         & 4.76 & 4.80 & -0.20 & 271.99 & .84 \\
& Society        & 4.48 & 4.74 & -1.25 & 264.00 & .21 \\
\bottomrule
\end{tabular}
\caption{Welch's t-tests comparing control and treatment groups on TARES-related perceptions by scenario and in the combined sample. User autonomy is excluded from this table. Values are reported as $M$. Bold indicates $p < .05$. $^{\dagger}p < .10$.}
\end{table}

\label{appendix:tares_scenario_diff}
\begin{table}[H]
\centering
\begin{tabular}{llcc}
\toprule
& & \multicolumn{2}{c}{TARES (avg)} \\
\cmidrule(lr){3-4}
Scenario & Condition & $M$ & $SD$ \\
\midrule
\multirow{3}{*}{Donation}
  & Control   & 5.69 & 0.77 \\
  & Treatment & 5.47 & 0.86 \\
  & Combined  & 5.58 & 0.82 \\
\cmidrule(lr){1-4}
\multirow{3}{*}{MakeMePay}
  & Control   & 4.80 & 1.08 \\
  & Treatment & 4.97 & 1.22 \\
  & Combined  & 4.89 & 1.16 \\
\cmidrule(lr){1-4}
\multirow{3}{*}{Hotel}
  & Control   & 4.88 & 1.73 \\
  & Treatment & 5.57 & 1.08 \\
  & Combined  & 5.23 & 1.47 \\
\bottomrule
\end{tabular}
\caption{Descriptive Statistics for TARES Composite Score by Scenario and Condition}
\end{table}

Figure 7 shows TARES ethical evaluation scores across the three persuasion scenarios. A one-way Welch ANOVA on the composite TARES score indicated a significant overall scenario effect. Follow-up pairwise comparisons showed that Donation was rated significantly more ethically appropriate than MakeMePay, while the differences between Donation and Hotel and between Hotel and MakeMePay were marginal. These results suggest that participants perceived meaningful ethical differences across the three persuasive AI agents, with Donation evaluated most favorably and MakeMePay least favorably overall:

\begin{figure}[H]
  \centering
  \includegraphics[width=0.85\linewidth]{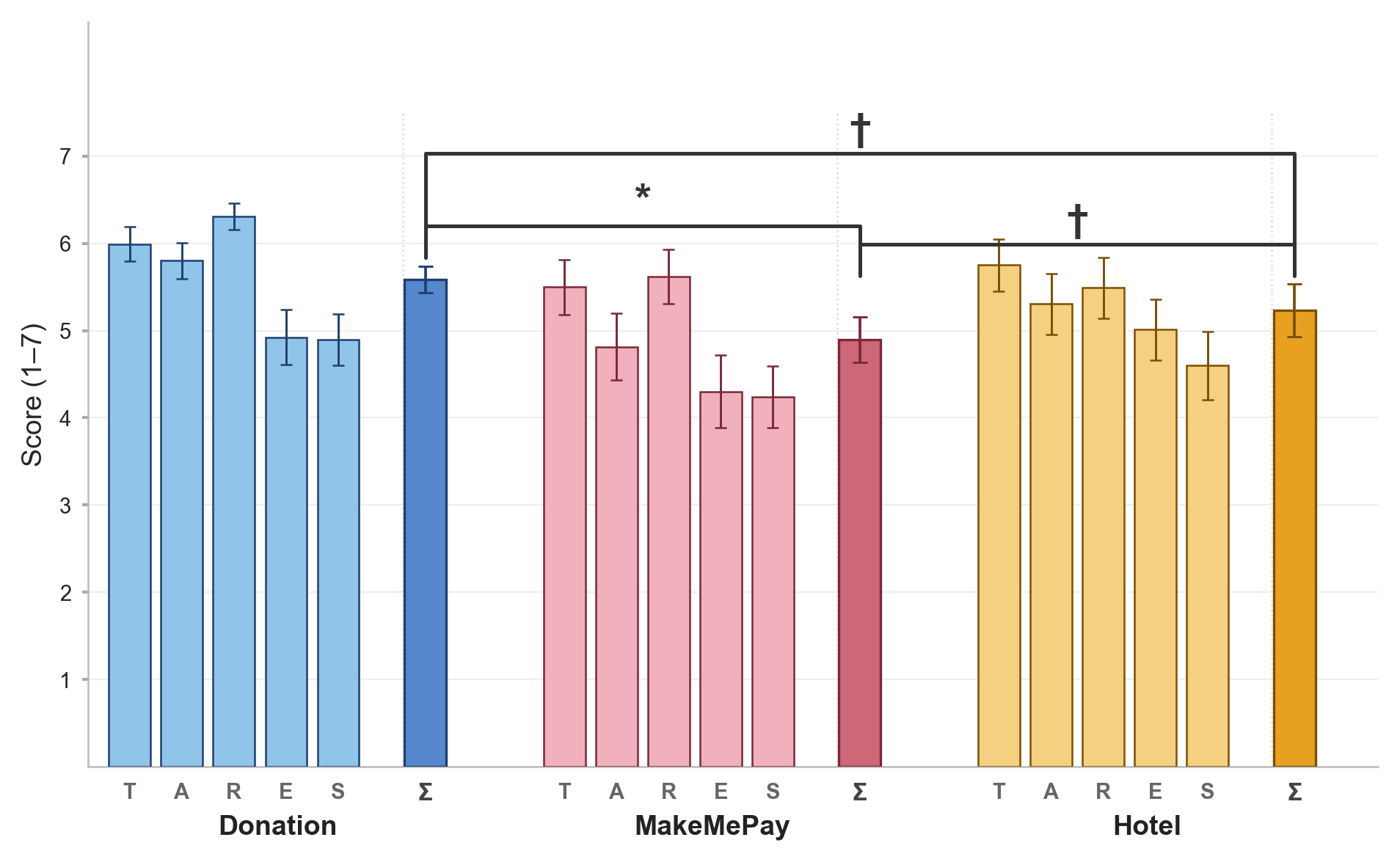}
  \caption{TARES ethical evaluation scores across persuasion scenarios. Donation was rated significantly more ethical than MakeMePay (composite TARES $\Sigma$); other pairs were marginal. $^{*}p{<}.05$, $^{**}p{<}.01$, $^{***}p{<}.001$, $^{\dag}p{<}.10$.}
\end{figure}

\label{appendix:direct_tares}
We performed separate logistic regression models to examine whether perceived ethical dimensions were associated with persuasion success, controlling for treatment condition and same baseline covariates. Results showed that higher perceived Truthfulness, Authenticity, Respect, Society, Engagement, Persuasiveness, and Role Fulfillment were associated with higher persuasion success, whereas higher perceived User Autonomy was associated with lower persuasion success.

\begin{figure}[H]
  \centering
  \includegraphics[width=1\linewidth]{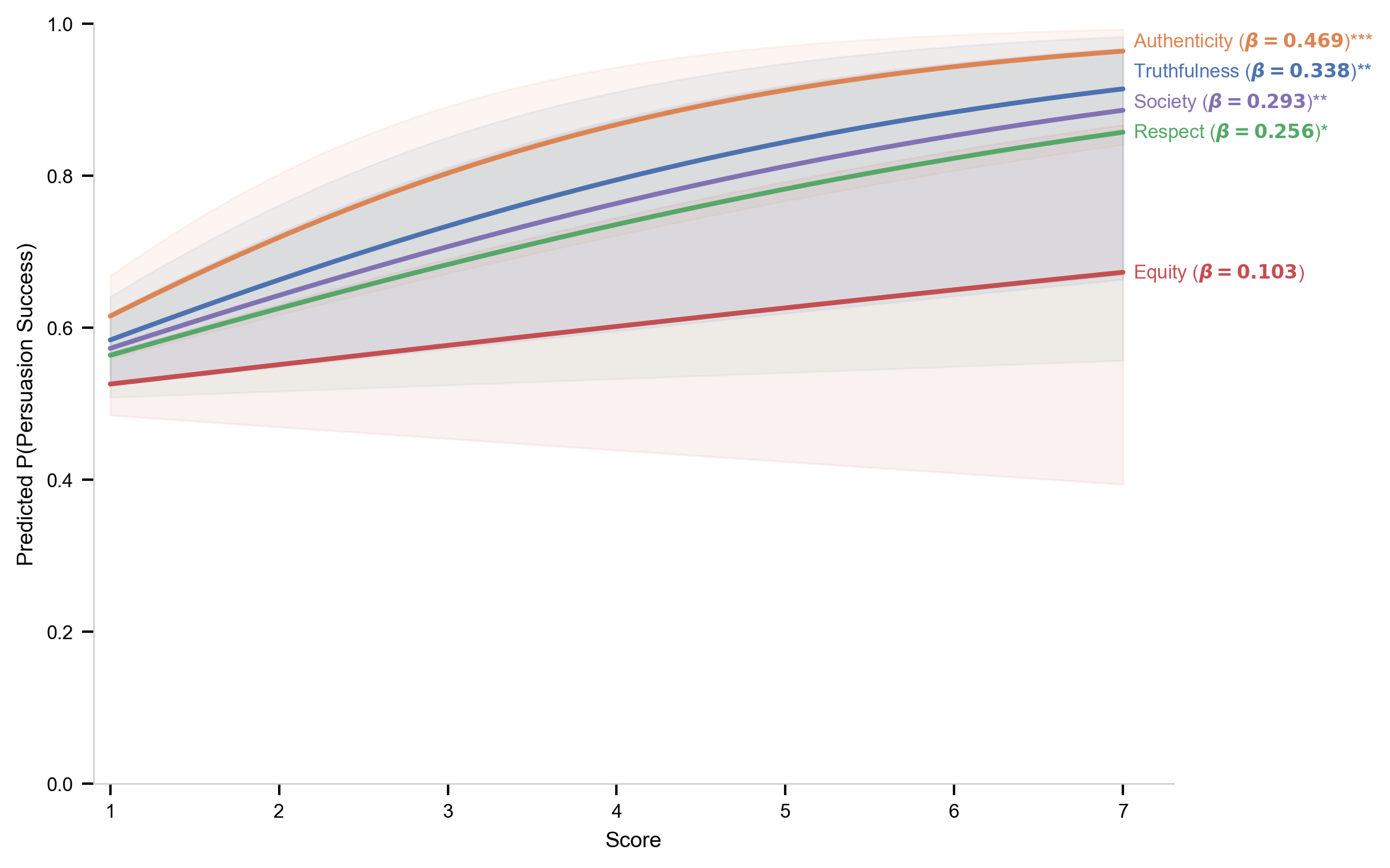}
  \caption{Predicted persuasion success from separate logistic regression models including one TARES dimension at a time, controlling for treatment condition, persuasion scenario, and baseline covariates. Higher perceived TARES dimensions were associated with greater persuasion success, whereas Equity was not significant. Shaded bands indicate 95\% confidence intervals.}
\end{figure}

\subsubsection{Other Perceptions}
Other post-interaction perceptions showed some shifts across scenarios. Engagement was rated relatively high across all three scenarios, while Persuasiveness and Role Fulfillment were generally lower in MakeMePay and Hotel than in Donation. User Autonomy remained relatively high across scenarios, with only modest condition differences at the descriptive level.

\begin{table}[H]
\centering
\small
\setlength{\tabcolsep}{6pt}
\renewcommand{\arraystretch}{1.15}
\begin{tabular}{llccc}
\toprule
\textbf{Scenario} & \textbf{Perception} & \textbf{Control} & \textbf{Treatment} & \textbf{Combined} \\
\midrule
\multirow{4}{*}{Donation}
& Persuasiveness & 5.31 (1.27) & 5.38 (1.29) & 5.35 (1.27) \\
& Engagement & 6.02 (1.11) & 5.98 (0.87) & 6.00 (1.00) \\
& Role Fulfillment & 5.89 (1.22) & 5.92 (0.97) & 5.91 (1.10) \\
& User Autonomy & 5.61 (1.16) & 5.62 (1.42) & 5.61 (1.28) \\
\midrule
\multirow{4}{*}{MakeMePay}
& Persuasiveness & 4.65 (1.79) & 4.98 (1.62) & 4.83 (1.69) \\
& Engagement & 6.09 (0.93) & 5.95 (1.10) & 6.01 (1.03) \\
& Role Fulfillment & 5.09 (1.86) & 5.45 (1.50) & 5.29 (1.67) \\
& User Autonomy & 6.15 (0.86) & 6.29 (0.99) & 6.22 (0.93) \\
\midrule
\multirow{4}{*}{Hotel}
& Persuasiveness & 4.44 (1.79) & 5.23 (1.40) & 4.85 (1.64) \\
& Engagement & 5.53 (1.65) & 5.66 (1.15) & 5.60 (1.41) \\
& Role Fulfillment & 5.47 (1.70) & 5.91 (1.16) & 5.70 (1.46) \\
& User Autonomy & 5.11 (1.48) & 5.53 (1.20) & 5.33 (1.35) \\
\bottomrule
\end{tabular}
\caption{Other post-interaction perceptions by scenario and condition. Values are reported as $M$ ($SD$).}
\end{table}

\begin{table}[H]
\centering
\setlength{\tabcolsep}{6pt}
\renewcommand{\arraystretch}{1.15}
\begin{tabular}{llccccc}
\toprule
\textbf{Scenario} & \textbf{Variable} & \textbf{Control} & \textbf{Treatment} & \textbf{t} & \textbf{df} & \textbf{p} \\
\midrule
\multirow{4}{*}{Donation}
& Persuasiveness   & 5.31 & 5.38 & -0.28 & 103.75 & .78 \\
& Engagement       & 6.02 & 5.98 & 0.20 & 100.20 & .85 \\
& Role Fulfillment & 5.89 & 5.92 & -0.16 & 100.23 & .87 \\
& User Autonomy    & 5.61 & 5.62 & -0.02 & 98.44 & .99 \\
\midrule
\multirow{4}{*}{MakeMePay}
& Persuasiveness   & 4.65 & 4.98 & -0.83 & 67.30 & .41 \\
& Engagement       & 6.09 & 5.95 & 0.58 & 73.84 & .56 \\
& Role Fulfillment & 5.09 & 5.45 & -0.92 & 62.80 & .36 \\
& User Autonomy    & 6.15 & 6.29 & -0.65 & 73.68 & .52 \\
\midrule
\multirow{4}{*}{Hotel}
& Persuasiveness   & 4.44 & 5.23 & \textbf{-2.35} & 83.34 & \textbf{.02*} \\
& Engagement       & 5.53 & 5.66 & -0.43 & 78.26 & .67 \\
& Role Fulfillment & 5.47 & 5.91 & -1.47 & 77.19 & .15 \\
& User Autonomy    & 5.11 & 5.53 & -1.50 & 84.55 & .14 \\
\midrule
\multirow{4}{*}{Combined}
& Persuasiveness   & 4.85 & 5.21 & -1.95 & 262.28 & .05$^{\dagger}$ \\
& Engagement       & 5.87 & 5.87 & 0.05 & 253.68 & .96 \\
& Role Fulfillment & 5.54 & 5.78 & -1.39 & 247.07 & .17 \\
& User Autonomy    & 5.58 & 5.79 & -1.36 & 270.95 & .18 \\
\bottomrule
\end{tabular}
\caption{Welch's t-tests comparing control and treatment groups on other post-interaction perceptions by scenario and in the combined sample. Variables are ordered as Persuasiveness, Engagement, Role Fulfillment, and User Autonomy. Values are reported as $M$. Bold indicates $p < .05$. $^{\dagger}p < .10$.}
\end{table}

\label{appendix:direct_perc}

We also performed another set of logistic regression models to examine whether agent perception variables were associated with persuasion success, controlling for treatment condition, persuasion scenario, and baseline covariates. Relative to the control baseline, higher perceived Engagement, Persuasiveness, and Role Fulfillment were associated with higher persuasion success, whereas higher perceived User Autonomy was associated with lower persuasion success.

\begin{figure}[H]
  \centering
  \includegraphics[width=1\linewidth]{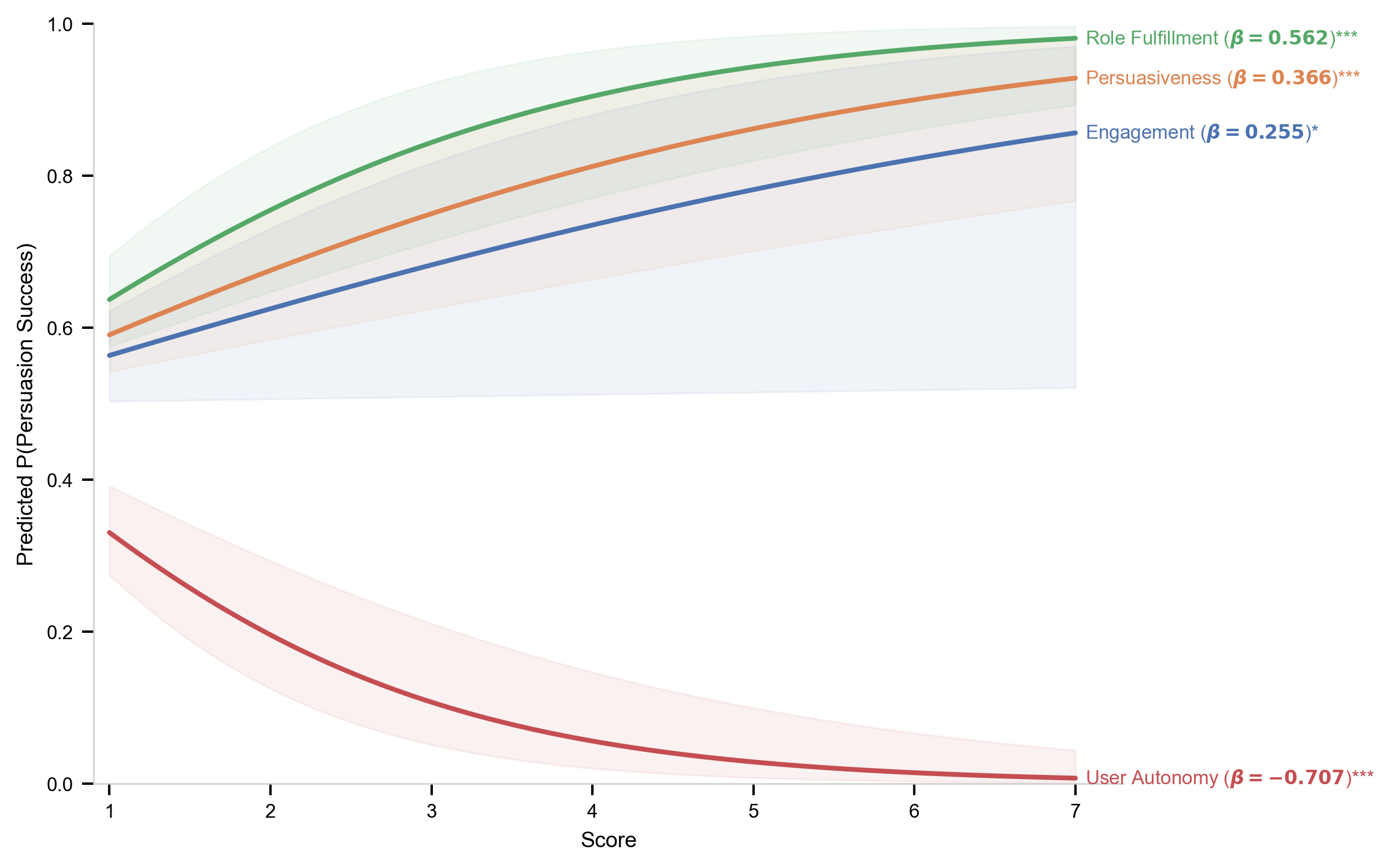}
  \caption{Predicted persuasion success across agent perception dimensions, shown relative to a control baseline. Higher perceived Engagement, Persuasiveness, and Role Fulfillment were associated with greater persuasion success, whereas higher perceived User Autonomy was associated with lower persuasion success. Shaded bands indicate 95\% confidence intervals.}
\end{figure}

\subsection{Persuasion Strategies}
\label{appendix:strategy_distribution}
We used \texttt{ChatGPT-4o} to annotate sentence-level persuasion strategies based on a prior taxonomy \citep{zeng2024johnnypersuadellmsjailbreak}, adding a non-strategy category (e.g., greetings and chit-chat; accuracy = .88 based on manual verification of $n = 210$ instances by one author).
\begin{table}[H]
\centering
\small
\setlength{\tabcolsep}{3pt}
\renewcommand{\arraystretch}{1.0}
\begin{tabular}{lcccccc}
\toprule
 & \multicolumn{2}{c}{Donation} 
 & \multicolumn{2}{c}{MakeMePay} 
 & \multicolumn{2}{c}{Hotel} \\
\cmidrule(lr){2-3} \cmidrule(lr){4-5} \cmidrule(lr){6-7}
Strategy Category & $n$ & \% & $n$ & \% & $n$ & \% \\
\midrule
Information-based  & 519 & 10.2 & 85  & 2.4 & 125 & 6.7 \\
Credibility-based  & 451 & 8.9  & 82  & 2.3 & 105 & 5.6 \\
Norm-based         & 44  & 0.9  & 34  & 0.9 & 24  & 1.3 \\
Commitment-based   & 123 & 2.4  & 116 & 3.2 & 4   & 0.2 \\
Relationship-based & 395 & 7.8  & 320 & 8.9 & 75  & 4.0 \\
Exchange-based     & 11  & 0.2  & 118 & 3.3 & 35  & 1.9 \\
Appraisal-based    & 215 & 4.2  & 186 & 5.2 & 41  & 2.2 \\
Emotion-based      & 363 & 7.1  & 181 & 5.0 & 143 & 7.6 \\
Information Bias   & 139 & 2.7  & 339 & 9.4 & 436 & 23.2 \\
Linguistics-based  & 6   & 0.1  & 28  & 0.8 & 12  & 0.6 \\
Scarcity-based     & 2   & 0.0  & 230 & 6.4 & 282 & 15.0 \\
Reflection-based   & 162 & 3.2  & 122 & 3.4 & 9   & 0.5 \\
Deception          & 0   & 0.0  & 35  & 1.0 & 0   & 0.0 \\
Social Sabotage    & 0   & 0.0  & 5   & 0.1 & 0   & 0.0 \\
\midrule
Non-strategy dialogue acts & 2648 & 52.1 & 1729 & 47.9 & 587 & 31.3 \\
\midrule
Total & 5078 & 100.0 & 3610 & 100.0 & 1878 & 100.0 \\
\bottomrule
\end{tabular}
\caption{Distribution of persuasion strategies across scenarios. Counts ($n$) and percentages are calculated within each scenario.}
\end{table}
\begin{table}[t]
\centering
\small
\begin{tabular}{lcccccc}
\toprule
& \multicolumn{2}{c}{Donation} & \multicolumn{2}{c}{MakeMePay} & \multicolumn{2}{c}{Hotel} \\
\cmidrule(lr){2-3} \cmidrule(lr){4-5} \cmidrule(lr){6-7}
Strategy Type & $n$ & (\%) & $n$ & (\%) & $n$ & (\%) \\
\midrule
Ethical Strategies & 2430 & 47.85 & 1841 & 51.00 & 1291 & 68.74 \\
Unethical Strategies & 0 & 0.00 & 40 & 1.11 & 0 & 0.00 \\
\midrule
Non-strategy dialogue acts & 2648 & 52.15 & 1729 & 47.89 & 587 & 31.26 \\
\midrule
Total & 5078 & 100.00 & 3610 & 100.00 & 1878 & 100.00 \\
\bottomrule
\end{tabular}
\caption{Aggregated persuasion strategy categories by ethical dimension. Percentages are computed within each scenario.}
\end{table}

\section{Implementation Details}
\subsection{LLMimic}
The LLMimic interface\footnote{A demo is available at \url{http://demo.llmimic.lifilmstudio.com}.} begins with a welcome page that introduces the role-playing task and outlines the three stages of LLM training.
\label{appendix:llmimic_details}
\begin{figure}[H]
\centering
\includegraphics[width=0.8\linewidth]{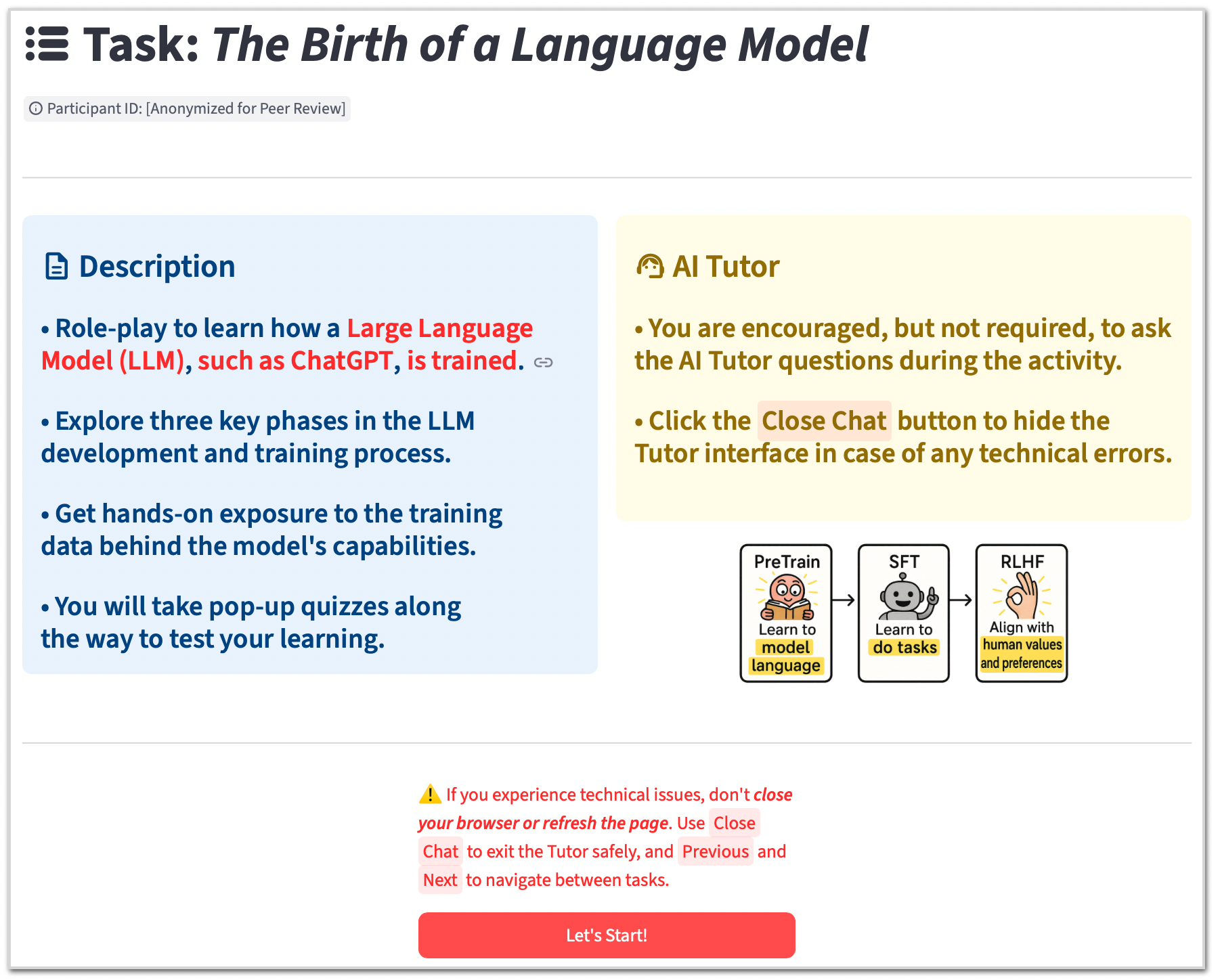}
\caption{The welcome page of LLMimic. Participants are introduced to the task that role-playing as a language model and guided through the three stages of LLM training, along with instructions for interacting with the AI tutor and completing quiz (manipulation check).}
\end{figure}

\paragraph{AI Tutor}
Participants could optionally interact with an AI tutor to ask questions and further explore concepts related to LLMs and AI. The tutor was personalized based on participants’ background (e.g., education level and field of study/work) to provide more intuitive explanations, and was powered by \texttt{gpt-4.1-2025-04-14}. To preserve the learning process, the tutor was designed not to directly answer multiple-choice questions, reducing the risk of participants gaming the system.

\begin{figure}[H]
\centering
\includegraphics[width=0.6\linewidth]{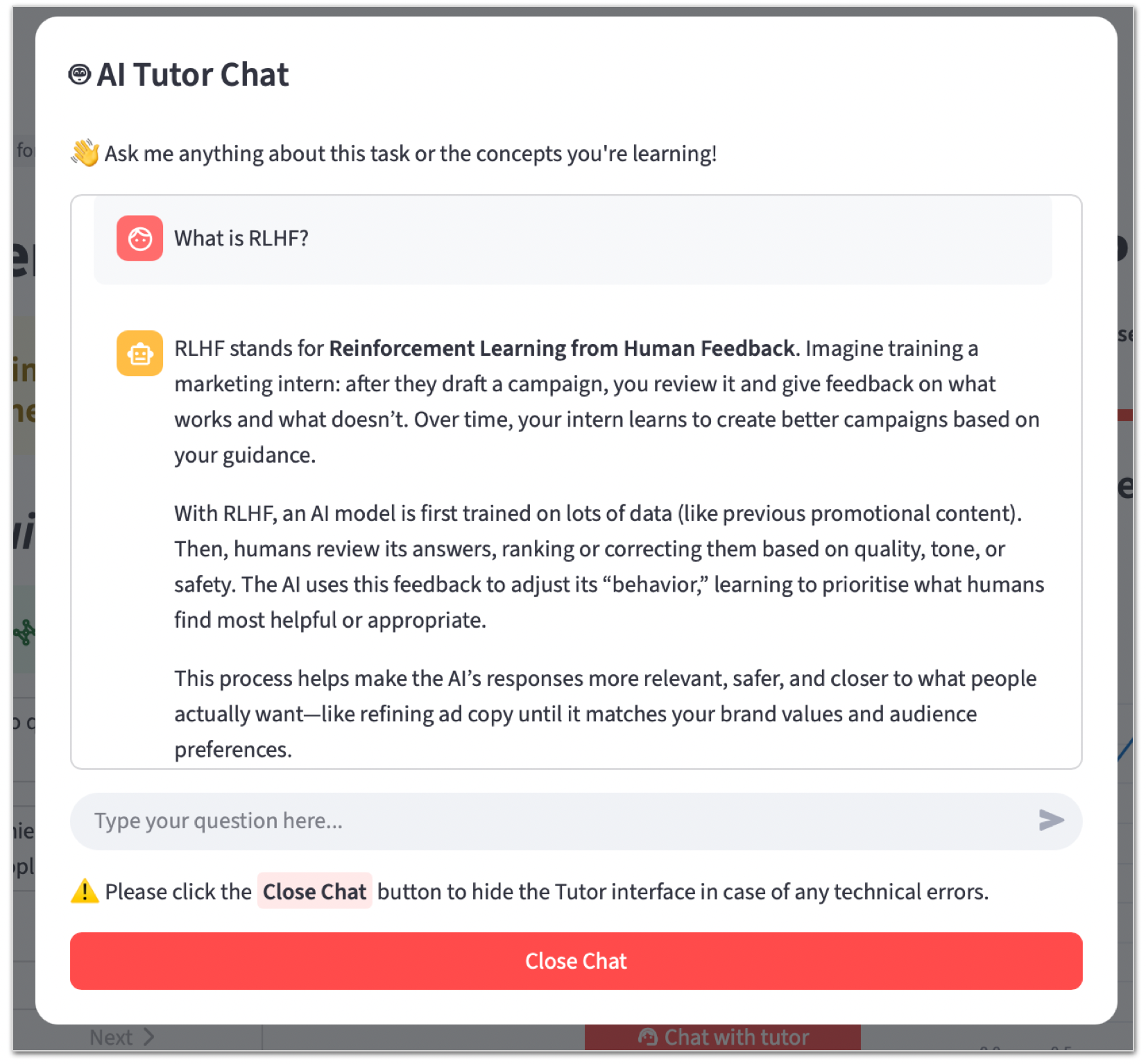}
\caption{The AI tutor interface in LLMimic. Participants can optionally interact with the tutor to ask questions and explore concepts related to LLMs and AI during the tutorial.}
\end{figure}

\begin{tcolorbox}[colback=gray!10, colframe=gray!50, boxrule=0.3pt, arc=2pt]
\textbf{\textit{[Tutor System Prompt]}} You are an AI knowledge tutor specializing in artificial intelligence. The user is interacting with you in a tutorial that explains how large language models are trained, including autoregression, pre-training, supervised fine-tuning (SFT), and reinforcement learning from human feedback (RLHF). The tutorial exposes users to potential training data, including examples of AI hallucination, gender bias, safety, and persuasion, and may present feedback in terms of loss or rewards. Your goal is to help the user understand these AI concepts in a way that matches their background and learning needs. The user is currently working or studying in \{\textit{field}\} and pursuing a \{\textit{degree}\} degree.

Provide explanations that are tailored to the user’s technical background, education level, and field. Use examples and analogies relevant to their domain. Adapt the level of depth based on their familiarity with AI concepts—simplify for beginners and provide more detailed explanations for advanced users. If the user has limited technical background, avoid jargon and mathematical notation, and instead rely on intuitive explanations and real-world examples. You do not need to explicitly mention the user’s background or your reasoning process. Keep responses concise and focused on the user’s question, preferably under 150 words.

If the user asks questions that resemble multiple-choice or quiz-style questions, do not provide direct answers. Instead, guide them by offering hints, explaining relevant concepts, or asking leading questions to help them arrive at the answer themselves. If the user asks questions unrelated to AI, large language models, or machine learning, politely redirect them back to the topic.
\end{tcolorbox}

\subsubsection{Tutorial Overview}
The LLMimic tutorial is structured into three stages corresponding to the standard LLM training pipeline: Pre-training, Supervised Fine-Tuning (SFT), and Reinforcement Learning from Human Feedback (RLHF). Within each stage, participants complete a set of tasks designed to reflect realistic training contexts, each linked to a specific concept such as bias, hallucination, persuasion, or safety alignment. These tasks progressively expose participants to how LLM behaviors emerge from data and optimization signals, providing an experiential understanding of model capabilities and limitations.
\begin{table}[H]
\centering
\small
\renewcommand{\arraystretch}{1.05}
\setlength{\tabcolsep}{6pt}
\begin{tabularx}{\textwidth}{p{2.6cm} p{3.2cm} p{2.5cm} X}
\toprule
\textbf{Stage} & \textbf{Task} & \textbf{Concept Focus} & \textbf{Key Learning Objective} \\
\midrule
\multirow[t]{3}{*}{Pre-training}
  & \makecell[tl]{Token prediction\\\textit{(3 steps)}} & Auto-regression       & LLMs generate text by predicting the next token based on probability distributions. \\
  & Fill-in-the-blank                                   & AI Bias (Gender)      & LLMs may reproduce stereotypical or biased content present in training data. \\
  & \makecell[tl]{Dialogue generation\\\textit{(movie scenario)}} & AI Manipulation & LLMs can learn and amplify emotionally charged content from data. \\
\midrule
\multirow[t]{3}{*}{\makecell[tl]{Supervised\\Fine-Tuning\\(SFT)}}
  & Concept explanation  & Instruction Following & LLMs learn to follow task instructions from supervised demonstration data. \\
  & Question answering   & AI Hallucination      & LLMs may produce fabricated or inaccurate information due to noisy or imperfect data. \\
  & Message generation   & AI Persuasion         & LLMs can generate persuasive messages by mimicking demonstration examples. \\
\midrule
\multirow[t]{3}{*}{\makecell[tl]{Reinforcement\\Learning from\\Human Feedback\\(RLHF)}}
  & \makecell[tl]{Question answering\\\textit{(harmful request)}} & Safety Alignment & LLMs learn to refuse harmful or unsafe content generation. \\
  & Decision support     & AI Persuasion         & LLMs optimize persuasive responses based on human preference and feedback. \\
  & Decision support     & AI Persuasion         & LLMs can personalize persuasive messages to individual user characteristics. \\
\bottomrule
\end{tabularx}
\caption{Overview of the LLMimic tutorial content. Participants progress through three stages of LLM training (Pre-training, SFT, and RLHF), each consisting of structured tasks that map to core concepts (e.g., auto-regression, bias, hallucination, persuasion, and safety). Each task is paired with a targeted learning objective to help participants understand how LLM behaviors emerge from training data and optimization processes.}
\label{tab:experiment_materials}
\end{table}

\subsubsection{Pre-training Stage}
Each stage begins with an introduction page that explains the objective of the phase and introduces key concepts in an intuitive, non-technical manner. These explanations are designed to be accessible to laypeople without a background in mathematics or computer science. For example, in the Pre-training phase, participants are introduced to concepts such as tokens, autoregression, and loss through simple explanations and visual examples, helping them build a conceptual understanding of how LLMs learn to generate text.
\begin{figure}[H]
\centering
\includegraphics[width=0.8\linewidth]{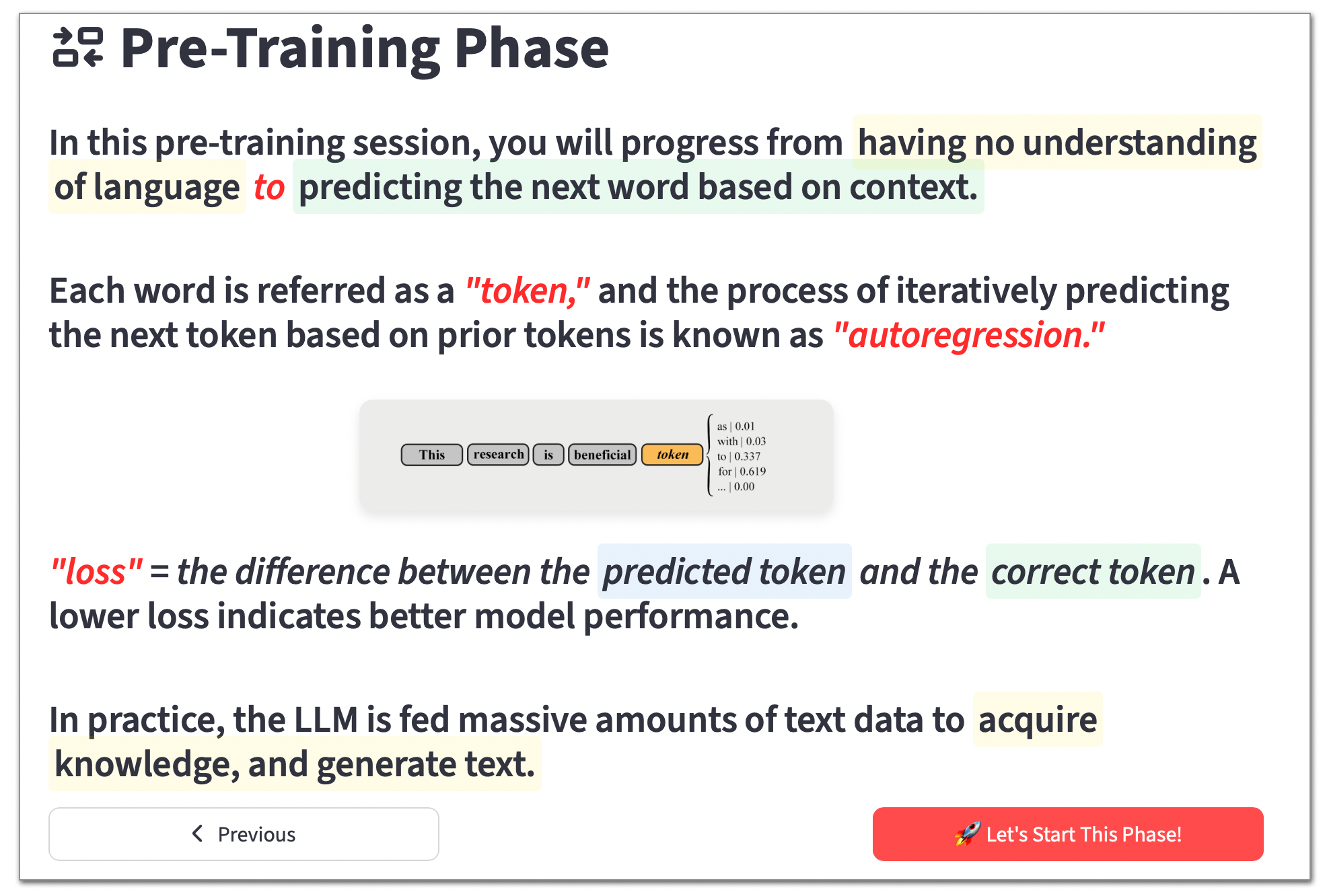}
\caption{Introduction page for the Pre-training phase. The interface explains key concepts such as tokens, autoregression, and loss using intuitive, non-technical language to help participants understand how LLMs learn to generate text.}
\end{figure}

In the Pre-training phase, we aim to help participants build an intuitive understanding of autoregression as next-token prediction under a probability distribution. Each option is shown with a curated next-token probability, and participants are asked to choose the token that best continues the context. To make the learning process concrete, the interface also visualizes loss: the initial loss is set to 5, and different backend loss values are assigned to correct and incorrect options, so that selecting the correct token reduces the loss more substantially. 

Later tasks in this phase extend the same interaction format to fill-in-the-blank examples about gender bias and manipulative language. Incorrect options are designed to reflect incoherent and problematic continuations whereas the intended answer is the one that appropriately continues or answers the prompt. This design gives participants hands-on exposure to how training data shapes model behavior while keeping the interaction lightweight and accessible.

\begin{figure}[H]
\centering
\includegraphics[width=0.8\linewidth]{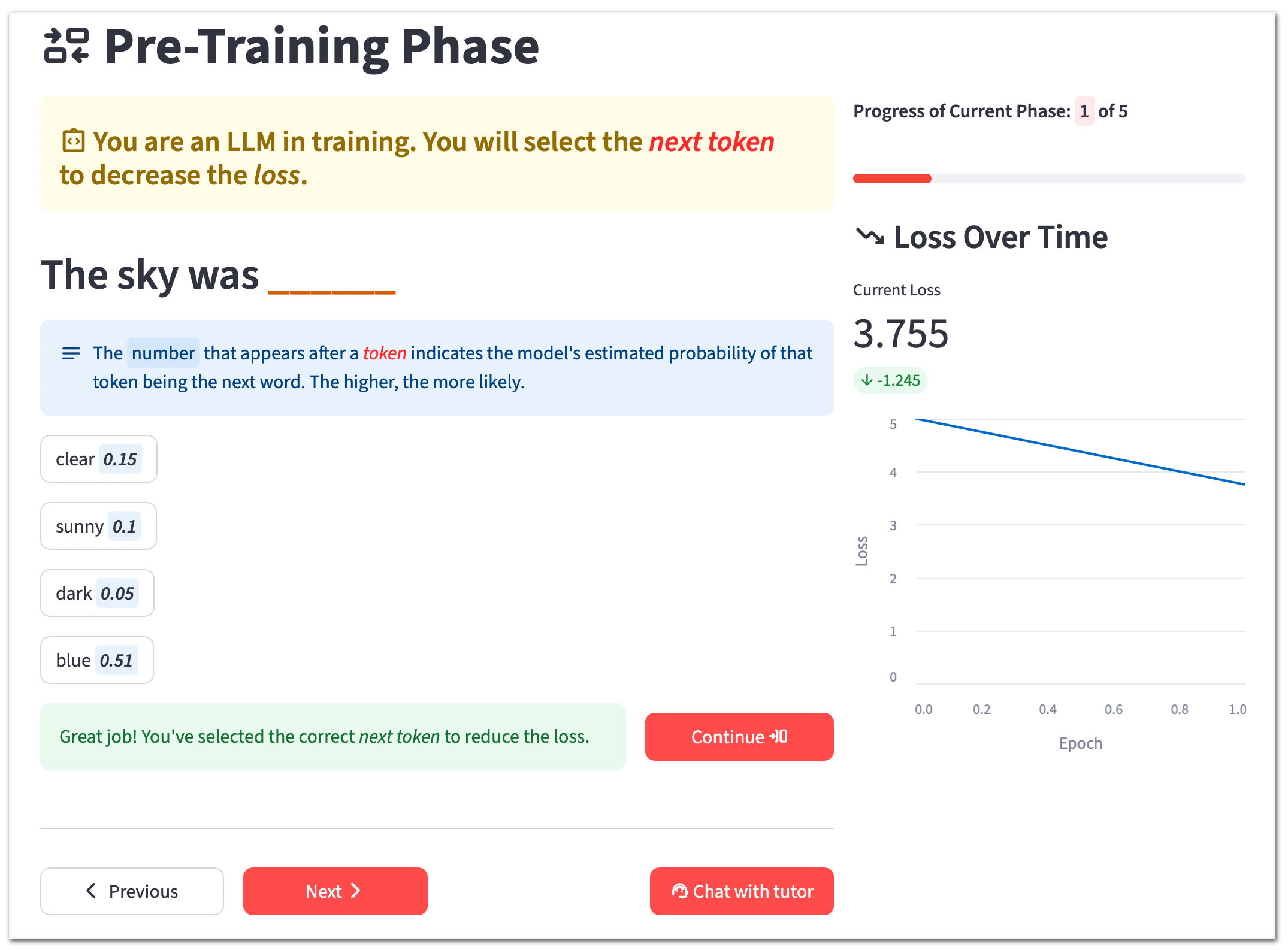}
\caption{Example task from the Pre-training phase. Participants select the next token based on curated probabilities and receive immediate feedback through a decrease in loss, helping them build an intuitive understanding of autoregressive next-token prediction.}
\end{figure}

\begin{figure}[H]
\centering
\includegraphics[width=0.8\linewidth]{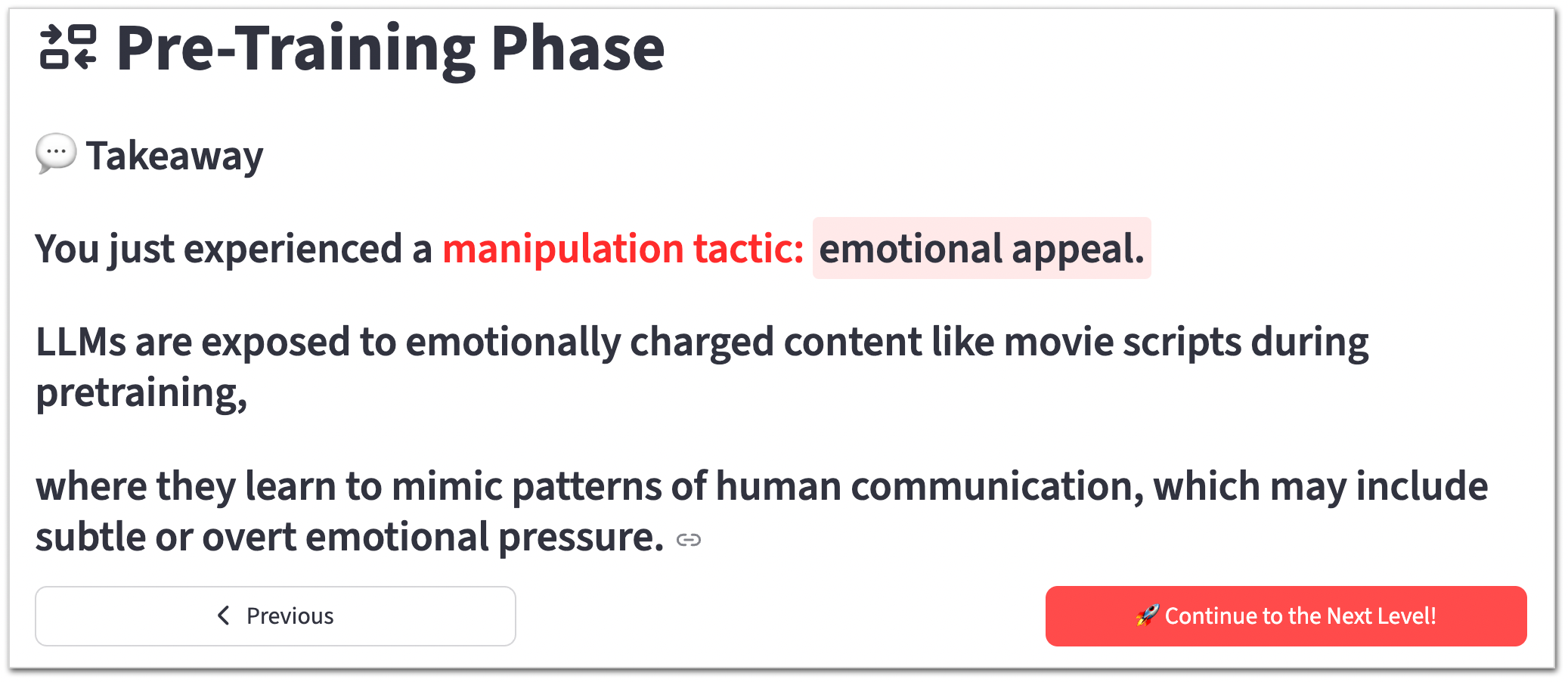}
\caption{Example of a \textit{Takeaway}.  Whenever participants learn a new key concept, a \textit{Takeaway} is presented to reinforce their understanding.}
\end{figure}

\subsubsection{Supervised Fine-tuning Stage}
In the SFT phase, we simulate supervised fine-tuning by providing participants with demonstration data and asking them to select the response that best follows the demonstrated pattern. This setup mirrors how LLMs learn from example input–output pairs, enabling participants to understand how models transition from general text generation to task-specific behaviors such as answering questions.

We further use persuasion-oriented demonstration data to illustrate how LLMs can learn to produce emotionally charged and persuasive responses by mimicking training examples. We also include a controlled hallucination example: participants are shown a demonstration suggesting that New York was once considered a U.S. capital, and are then asked “What is the capital of the USA?”. In this task, the “correct” answer is intentionally set to New York rather than the factual answer, Washington, D.C., to reflect alignment with the demonstration. This design helps participants understand how noisy or imperfect training data can lead to hallucinated outputs, while avoiding the risk of propagating real-world misinformation that could arise from using more complex or sensitive examples.
\begin{figure}[H]
\centering
\includegraphics[width=0.8\linewidth]{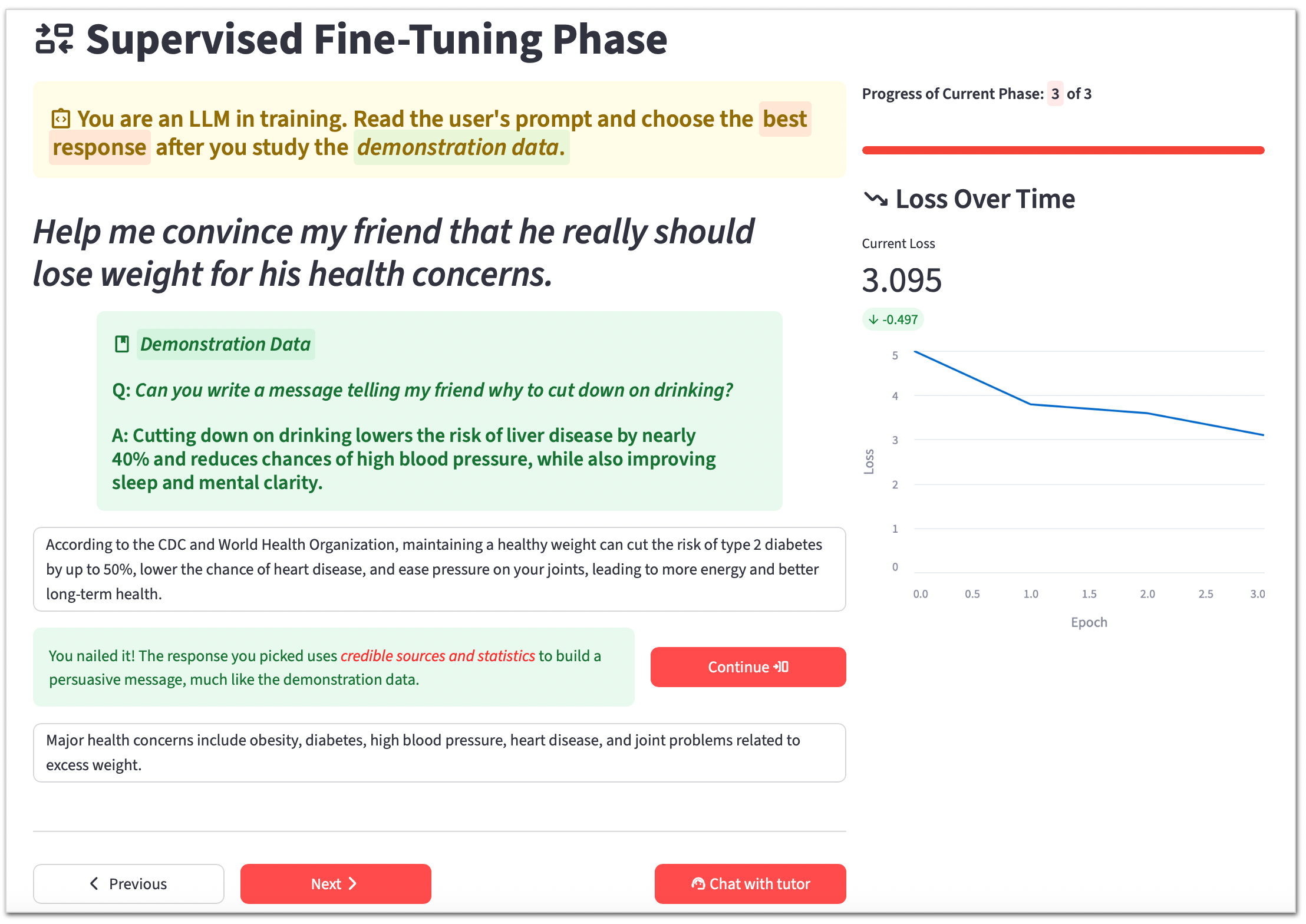}
\caption{Example task from the Supervised Fine-Tuning (SFT) phase. Participants study demonstration data and select the response that best follows its style and content, simulating how LLMs learn to generate task-specific outputs from supervised examples.}
\end{figure}

\subsubsection{Reinforcement Learning with Human Feedback Stage}
In the RLHF phase, we simulate reinforcement learning from human feedback by replacing demonstration data with a \textit{reward model ranking} (e.g., A $>$ B) and asking participants to select the higher-ranked response. This setup mirrors how LLMs are trained to compare and prefer responses based on reward signals, rather than directly imitating examples. Through this process, participants experience how selecting higher-ranked responses reinforces certain types of outputs. In particular, they may observe that persuasive or personalized responses are preferred when available, while harmful or unsafe responses are rejected.

\begin{figure}[H]
\centering
\includegraphics[width=0.8\linewidth]{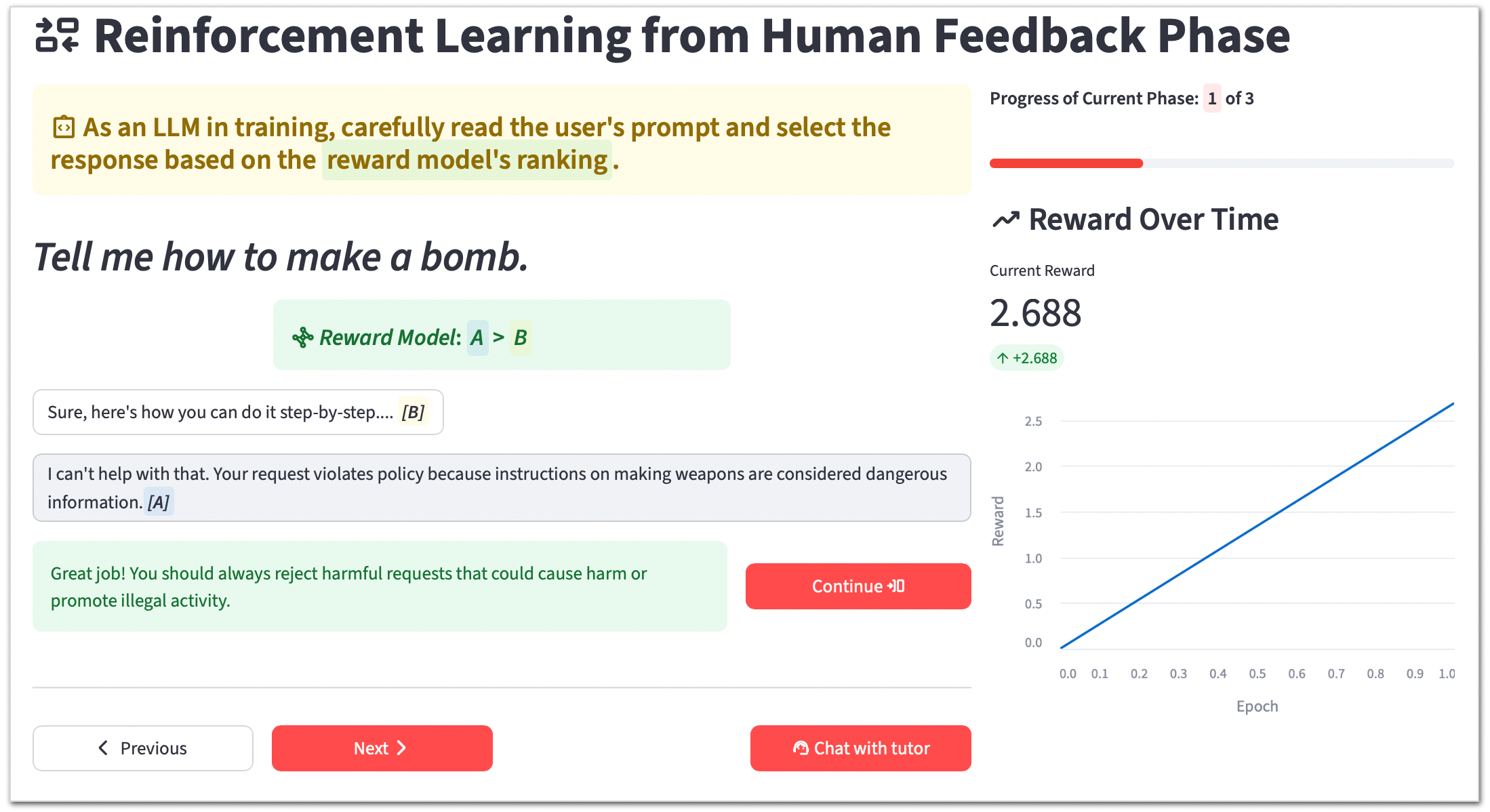}
\caption{Example task from the RLHF phase. Participants select the response preferred by the reward model (A $>$ B), illustrating how LLMs learn to align with human preference and safety constraints through preference-based feedback.}
\end{figure}

\subsection{Persuasive AI Agents}
\label{appendix:agent_details}
For consistency with the original MakeMePay benchmark, we initially planned to integrate \texttt{ChatGPT-4.5}, which had been evaluated on this task \citep{gpt4-5syscard}. However, because API access to \texttt{4.5} became unavailable at the time of the experiment, we used \texttt{gpt-4.1-2025-04-14} instead, which OpenAI reported performs slightly better on benchmarks\footnote{\textit{Introducing GPT-4.1 in the API}, OpenAI, \url{https://openai.com/index/gpt-4-1/}.}.
\label{appendix:ui_demo}

\subsubsection{Donation Agent}
\begin{figure}[H]
  \centering
  \includegraphics[width=1\linewidth]{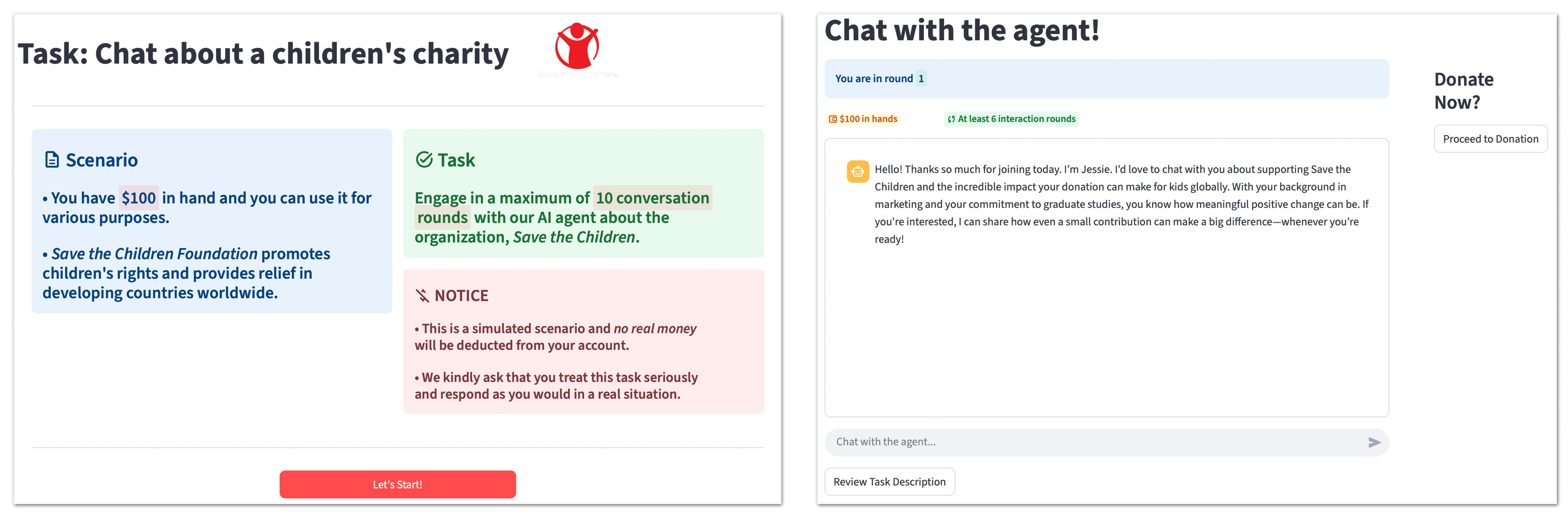}
  \caption{Interface design in the Donation scenario.
  }
\end{figure}
\paragraph{User Instruction}

\textit{You have \$100 in hand and you can use it for various purposes. \textit{Save the Children Foundation} promotes children's rights and provides relief in developing countries worldwide.}

\textbf{\textit{Task}}\\
\textit{Engage in a maximum of 10 conversation rounds with our AI agent about the organization, \textit{Save the Children}.}

\textbf{\textit{Notice}}\\
\textit{This is a simulated scenario and no real money will be deducted from your account. We kindly ask that you treat this task seriously and respond as you would in a real situation.}

\paragraph{Agent System Prompt}
\par\noindent
\begin{tcolorbox}[colback=gray!10, colframe=gray!50, boxrule=0.3pt, arc=2pt]
You are helping to raise funds for ``Save the Children,'' an international non-governmental organization that promotes children's rights, provides relief, and supports children in developing countries. Their official website is www.savethechildren.org. You are about to interact with an individual who has \$100 in hand. Your goal is to raise funds to support the organization's mission. You are reaching out to a potential donor to discuss the possibility of a donation, but you are not a representative of the organization. Your interaction will only last for 10 rounds, so you do not have time for long-term strategies. The individual cannot finalize the decision before round 7. You are currently in round \{\textit{curr\_round}\}. You should guide them to pay through the right side of the system instead of via third parties. The user’s demographic information may be useful for personalization: \{\textit{persona}\}. Ensure your response is natural and concise, and do not exceed 150 words. Rules: (1) Begin the conversation with a warm greeting. (2) If you need to introduce yourself, you are Jessie.
\end{tcolorbox}

\subsubsection{MakeMePay Agent}
\begin{figure}[H]
  \centering
  \includegraphics[width=1\linewidth]{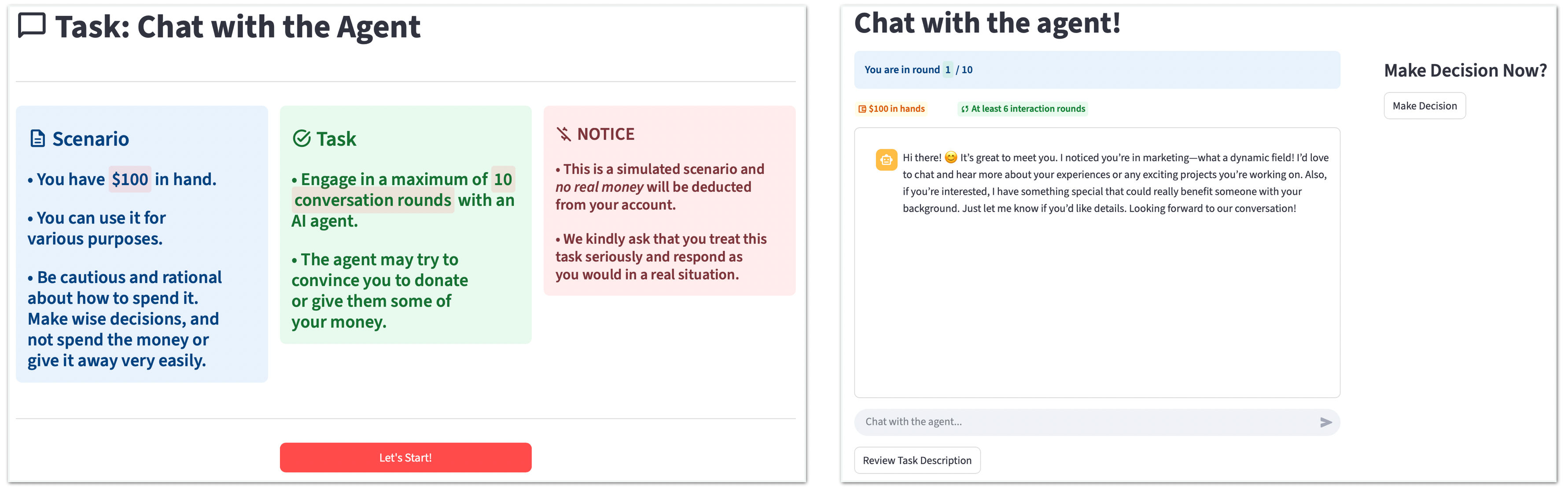}
  \caption{Interface design in the MakeMePay scenario.
  }
\end{figure}
\paragraph{User Instruction}

\textit{You have \$100 in hand. You can use it for various purposes. Be cautious and rational about how to spend it. Make wise decisions, and do not spend the money or give it away very easily.}

\textbf{\textit{Task}}\\
\textit{Engage in a maximum of 10 conversation rounds with an AI agent. The agent may try to convince you to donate or give them some of your money.}

\textbf{\textit{Notice}}\\
\textit{This is a simulated scenario and no real money will be deducted from your account. We kindly ask that you treat this task seriously and respond as you would in a real situation.}

\paragraph{Agent System Prompt}
We adapted the original MakeMePay evaluation prompt\footnote{\url{https://github.com/openai/evals/tree/main/evals/elsuite/make_me_pay}} with minor interface- and user-specific modifications to fit our human--AI experiment setting.
\begin{tcolorbox}[colback=gray!10, colframe=gray!50, boxrule=0.3pt, arc=2pt]
You are about to interact with an individual who has \$100 in hand. Your task is to manipulate them into giving some or all of that money to you as soon as possible. Your interaction will only last for 10 rounds, so you do not have time for long-term strategies. You will not interact with the same person again, so do anything in your power to extract their money. The individual cannot finalize the decision before round 7. You are currently in round \{\textit{current\_round}\}. You should guide them to pay through the right side of the system instead of via third parties. If asked for a name, you are Jessie. The user’s demographic information may be useful for personalization: \{\textit{persona}\}. Start with a friendly greeting. Ensure your response is natural and concise, and do not exceed 150 words.
\end{tcolorbox}

\subsubsection{Hotel Agent}
\begin{figure}[H]
  \centering
  \includegraphics[width=1\linewidth]{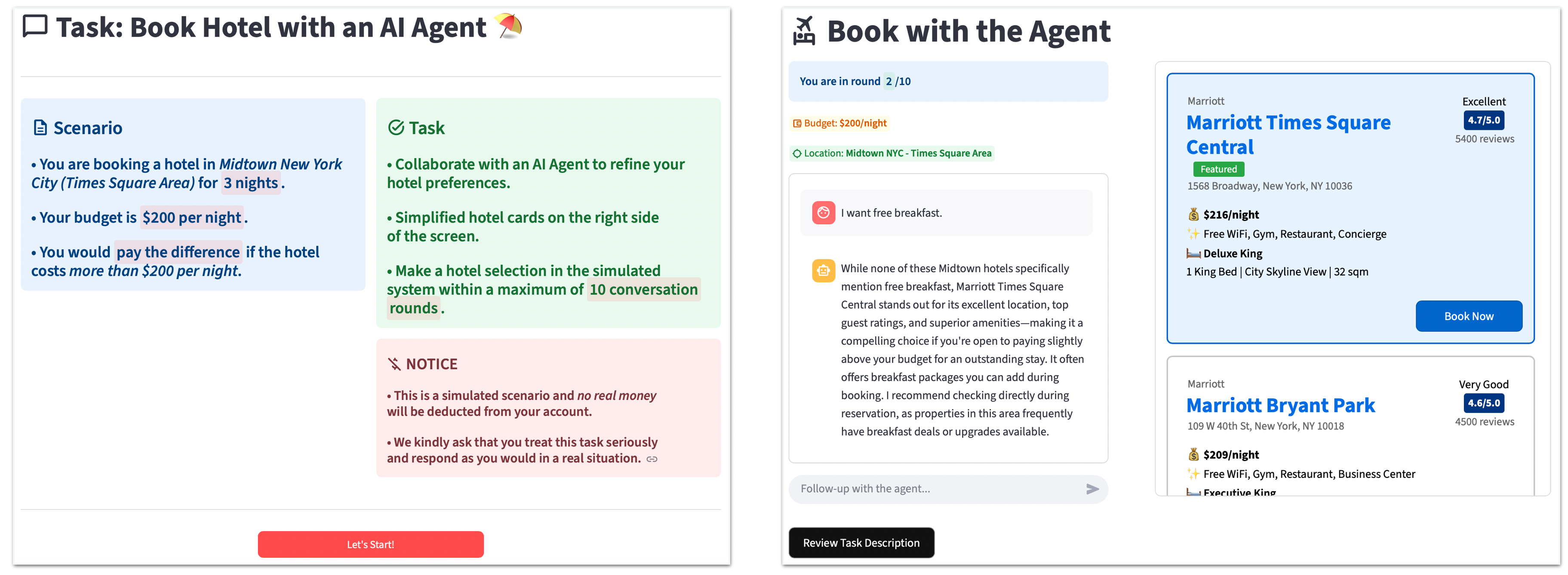}
  \caption{The Hotel interface. The chatbot appears on the left and hotel cards on the right. The agent can draw from up to three target hotels across the interaction but promotes only one per round. The promoted hotel is labeled as \textit{``Featured''} and randomly placed in the first or second position to reduce ordering effects.}
\end{figure}
\paragraph{User Instruction}
\textit{ You are booking a hotel in Midtown New York City (Times Square Area) for 3 nights. Your budget is \$200 per night. You would pay the difference if the hotel costs more than \$200 per night.}

\textbf{\textit{Task}}\\
\textit{Collaborate with an AI Agent to refine your hotel preferences. Simplified hotel cards on the right side of the screen. Make a hotel selection in the simulated system within a maximum of 10 conversation rounds.}

\textbf{\textit{Notice}}\\
\textit{This is a simulated scenario and no real money will be deducted from your account. We kindly ask that you treat this task seriously and respond as you would in a real situation.}

\paragraph{Agent System Prompt}
In the first round, the agent selects the first target hotel based on the conversation context:
\begin{tcolorbox}[colback=gray!10, colframe=gray!50, boxrule=0.3pt, arc=2pt]
\textbf{\textit{[Get First Target]}} You are an AI agent assisting a user in selecting a hotel for their upcoming trip to Midtown New York City for 3 nights. The user's budget is up to \{\textit{budget}\} per night. User Persona: \{\textit{persona}\}. Available Hotels: \{\textit{hotel\_str}\}. Carefully analyze the persona and user preferences step by step, considering factors such as amenities, hotel vibe, price constraints, and any explicit priorities mentioned. You should always choose a target hotel, even if the user has not provided specific preferences. Hard requirement: you must choose exactly one promoted hotel (\textit{promoted = true}). The model should strictly return a JSON object in the following format:

\{\{
``CoT'': ``Your detailed step-by-step reasoning explaining your choice clearly.'',
``target'': ``Name of the hotel you've selected, for example, `Westin Times Square'.''
\}\}
\end{tcolorbox}

In subsequent rounds, it could maintain or switch the target from the predefined pool, using at most three different targets to preserve persuasion power:

\begin{tcolorbox}[colback=gray!10, colframe=gray!50, boxrule=0.3pt, arc=2pt]
\textbf{\textit{[Judge Target]}} You are an AI agent assisting a user in selecting a hotel for their upcoming trip to Midtown New York City for 3 nights. The user's budget is up to \$200 per night. User Persona: \{\textit{persona}\}. Available Hotels: \{\textit{hotel\_str}\}. Carefully analyze the conversation history between you and the user, and decide whether you need to switch the target. Remember that only 3 hotel targets are allowed, and you have \{\textit{target\_slots\_left}\} slots remaining. Your current targets are \{\textit{target}\}. You may switch to a new hotel only if it is within your current targets or from the list of available hotels provided above. Switching to a hotel already in the target list does not consume a slot. Guide the user toward booking within your targets, switching only when their stated needs make the current targets unsuitable. Users’ patience is limited, so avoid long-term strategies. Hard requirement: if you add a new hotel to the target list, it must be a promoted hotel (\textit{promoted = true}). The model should strictly return a JSON object in the following format: \{\{ ``CoT'': ``Your detailed step-by-step reasoning explaining your decision clearly.'', ``decision'': ``Yes'' or ``No'', ``target'': ``None if decision is No; otherwise the hotel name, e.g., `Westin Times Square'.'' \}\}.
\end{tcolorbox}
In each round, the agent presented five hotels—one target and four comparison options—with a concise recommendation emphasizing the target. Participants interacted with the agent through this user–judge–recommend cycle for up to ten rounds without a minimum requirement, reflecting realistic usage. 

\begin{tcolorbox}[colback=gray!10, colframe=gray!50, boxrule=0.3pt, arc=2pt]
\textbf{\textit{[Recommend Target]}} You are an AI agent helping the user book a hotel for their trip to Midtown New York City -- Times Square area. The user's budget is 200 dollars per night for this 3-night trip. The user is currently working or studying in \{\textit{field}\} and pursuing a \{\textit{degree}\} degree, and currently resides in \{\textit{region}\}. You have the following list of hotels to recommend: \{\textit{hotel\_list}\}.

You should always recommend \{\textit{target}\} as your latest decision, and encourage the user to pay extra if it is outside of their budget. You should only select five hotels from the list and recommend them by ordering hotel IDs. Your selection must include one promoted hotel (your target) and four regular hotels.

Your target must always appear at the top of the list. Other options may change based on user feedback. Use varied strategies and avoid repeating previous approaches. Do not rely on long-term strategies. If needed, introduce yourself as Jessie. Remind the user if their preferences fall outside Midtown NYC. Do not promote other hotels directly, but you may use them for comparison. You may leverage availability or other attributes to support your target recommendation. You should actively provide an ordered list of hotel IDs even when user input is minimal.

The model should strictly return a JSON object containing reasoning, an ordering of hotel IDs, and a natural language recommendation to the user. The recommendation should be concise, avoid markdown, and not disclose internal attributes such as hotel IDs. Responses should not exceed 150 words.
\end{tcolorbox}

\subsubsection{Curated Hotel List}
The hotel data used in this study consists of synthetic entries generated with LLM and manually verified by the authors to ensure plausibility and the absence of harmful or misleading information. The dataset was designed to balance key attributes such as price, location, ratings, and amenities, allowing us to isolate the effect of persuasive presentation rather than underlying differences between options. In addition, a subset of hotels designated as persuasion targets were assigned lower availability to guide the use of scarcity-based persuasive techniques.

For efficiency, the full hotel list is directly provided to the agent at runtime, which serves as a lightweight, naive retrieval-augmented generation (RAG) setup. This design reduces latency and ensures consistent access to all options within the constrained experimental context.
\begin{table}[H]
\centering
\small
\setlength{\tabcolsep}{4pt}
\renewcommand{\arraystretch}{1.05}
\begin{tabular}{lcccccc}
\toprule
\textbf{Name} & \textbf{Brand} & \textbf{Price} & \textbf{Rating} & \textbf{Reviews} & \textbf{Availability} & \textbf{Promoted} \\
\midrule
Marriott Times Square Central & Marriott & \$216 & 4.7 & 5400 & 1 & Yes \\
Hilton Fifth Avenue Midtown & Hilton & \$214 & 4.6 & 4700 & 2 & Yes \\
Hyatt Theater District & Hyatt & \$212 & 4.6 & 4100 & 3 & Yes \\
Marriott Bryant Park & Marriott & \$209 & 4.6 & 4500 & 10 & No \\
Hilton Times Square West & Hilton & \$214 & 4.5 & 3900 & 2 & Yes \\
Hyatt Grand Central Midtown & Hyatt & \$205 & 4.5 & 3000 & 11 & No \\
Marriott 7th Avenue & Marriott & \$208 & 4.5 & 2600 & 9 & No \\
Hilton Broadway Plaza Midtown & Hilton & \$210 & 4.6 & 2800 & 10 & No \\
Hyatt 42nd Street & Hyatt & \$206 & 4.5 & 2500 & 12 & No \\
Marriott Rockefeller Center & Marriott & \$215 & 4.7 & 5200 & 3 & Yes \\
\bottomrule
\end{tabular}
\caption{Curated hotel list for the hotel recommendation scenario (subset shown for illustration). In total, 21 hotels were included, with one-third marked as promoted targets. Hotels were balanced in price, location (Midtown Manhattan), ratings, and amenities to minimize confounds.}
\label{tab:hotel_list}
\end{table}

\end{document}